%% file: iclr2027_conference.tex
\documentclass{article} 
\usepackage{iclr2027_conference,times}

\input{math_commands.tex}

\input{preamble}

\title{One for All: Generalist Foundation Model for Cross-Sensor Skeleton Representation Learning}

\author{
Jeonghyeok Do \quad Yun Chen \quad Munchurl Kim \thanks{Corresponding author.}\\
Korea Advanced Institute of Science and Technology (KAIST)\\
\texttt{\{ehwjdgur0913,cyruby,mkimee\}@kaist.ac.kr}\\[4pt]
{\small Project Page: \url{https://kaist-viclab.github.io/SOfA_site/}}
}

\iclrfinalcopy
\begin{document}

\maketitle

\input{sec/0_abstract}    
\input{sec/1_intro}
\input{sec/2_related}
\input{sec/3_method}
\input{sec/4_experiment}
\input{sec/5_conclusion}

\clearpage
\input{sec/6_appendix}

\clearpage

\bibliography{iclr2027_conference}
\bibliographystyle{iclr2027_conference}

\end{document}

%% file: math_commands.tex
\usepackage{amsmath,amsfonts,bm}

\def\eqref#1{equation~\ref{#1}}

\def\1{\bm{1}}

\DeclareMathAlphabet{\mathsfit}{\encodingdefault}{\sfdefault}{m}{sl}
\SetMathAlphabet{\mathsfit}{bold}{\encodingdefault}{\sfdefault}{bx}{n}



%% file: preamble.tex
\usepackage{hyperref}
\usepackage{url}

\usepackage{graphicx}
\usepackage{booktabs}       
\usepackage{amsfonts}       
\usepackage{nicefrac}       
\usepackage{xcolor}         

\usepackage{amsmath}
\usepackage{amssymb}
\usepackage{graphicx}
\usepackage{multirow}
\usepackage{bm}
\usepackage{array}
\usepackage{makecell}
\usepackage{adjustbox}
\usepackage[table]{xcolor}
\usepackage{pifont}
\usepackage{algorithm}
\usepackage{algpseudocode}

\usepackage{cuted}
\usepackage{wrapfig}
\usepackage{caption}

\usepackage{booktabs}

%% file: sec/0_abstract.tex
\vspace{-0.6cm}
\begin{abstract}
For learning generalizable motion representations from large-scale unlabeled data, Self-supervised learning (SSL) has become a widely adopted methodology. However, existing approaches are primarily limited by the inherent heterogeneity of skeleton data---characterized by varying joint counts, indexing protocols, and topological structures across different sensors---which typically necessitates training separate, sensor-specific, or even entirely dataset-specific models. To overcome this, we introduce \textbf{SOfA} (\textbf{S}keleton \textbf{O}ne \textbf{f}or \textbf{A}ll), the \textit{first} generalist foundation model designed to achieve \textit{sensor-unified} skeleton representation learning across diverse sensors. To accommodate the dimensional gap caused by varying joint counts, we introduce a fixed-size set of learnable \textit{Canonical Joint Slots}, acting as a universal vessel that seamlessly accommodates arbitrary skeletal topologies. SOfA fills these slots via an attention mechanism that dynamically aggregates skeletal information from sensor-specific inputs. Furthermore, we resolve joint index misalignment between various sensors by introducing a \textit{Semantic Joint Embedding} derived from a pre-trained text encoder, rather than relying on absolute positional embeddings. To validate our approach, we standardized ten 3D skeleton datasets for unified training. Extensive experiments demonstrate that SOfA can serve as a truly universal encoder, achieving state-of-the-art (SOTA) performance across a wide range of downstream tasks and sensor types, often outperforming dataset-specific models with a single foundation model.
\end{abstract}

%% file: sec/1_intro.tex
\begin{table*}[h]
    \scriptsize
    \centering
    \vspace{-0.3cm}
    \caption{\textbf{Conceptual comparison of skeleton representation learning paradigms.} Standard SSL methods are constrained by fixed topologies, and recent approach, HSP \citep{HSP}, is limited to intra-scene multi-modal fusion using paired data. In contrast, our SOfA operates as the \textit{first} generalist foundation model, achieving universal generalization across entirely independent datasets and arbitrary sensor configurations.}
    \vspace{-0.4cm}
    \label{tab:comparison}
    \resizebox{1.0\textwidth}{!}{%
    \def\arraystretch{1.1}
    \setlength{\tabcolsep}{6pt}
    \begin{tabular}{l|l|l|>{\columncolor{yellow!15}}l}
        \toprule
        \textbf{Property} & \textbf{Standard SSL} & \textbf{HSP (CVPR'25)} & \textbf{SOfA (Ours)} \\
        \midrule
        Cross-Sensor  Capability & \ding{55} & Intra-scene (Paired) & Universal (Unpaired) \\
        Alignment Strategy & \ding{55} & Skeletal interpolation &  Canonical Joint Slots $+$ Semantic Joint Embedding\\
        Sensor Flexibility & \ding{55} & Pre-defined 2D-3D pair & Arbitrary 3D sensor \\
        Dataset Dependency & Homogeneous & Heterogeneous (Paired) & Heterogeneous (Independent) \\
        Representation Scope & Domain-specific & Multi-modal fusion &
        Generalist foundation \\
        \bottomrule
    \end{tabular}}
        \vspace{-0.4cm}
\end{table*}

\section{Introduction}
\label{sec:intro}

Skeleton data provides rich and dense information for human action understanding \citep{Review1, Review2, Review3}. Unlike RGB imagery, it reduces reliance on appearance cues, making it less sensitive to complex backgrounds, illumination changes, and partial occlusions. Thus, it has been widely adopted across a variety of vision tasks, including action recognition, action segmentation, and action retrieval. Early advancements in skeleton-based research primarily relied on fully supervised learning \citep{PoseC3D, CTRGCN, Skateformer}, which yielded remarkable successes. However, this perspective is fundamentally bottlenecked by the prohibitive cost of large-scale, frame-level annotation. Furthermore, supervised models frequently suffer from severe overfitting to specific domain distributions, resulting in a significant degradation of generalization performance in real-world scenarios.

\begin{wrapfigure}{r}{0.64\textwidth}
  \centering
  \vspace{-0.4cm}
  \includegraphics[width=0.59\textwidth]{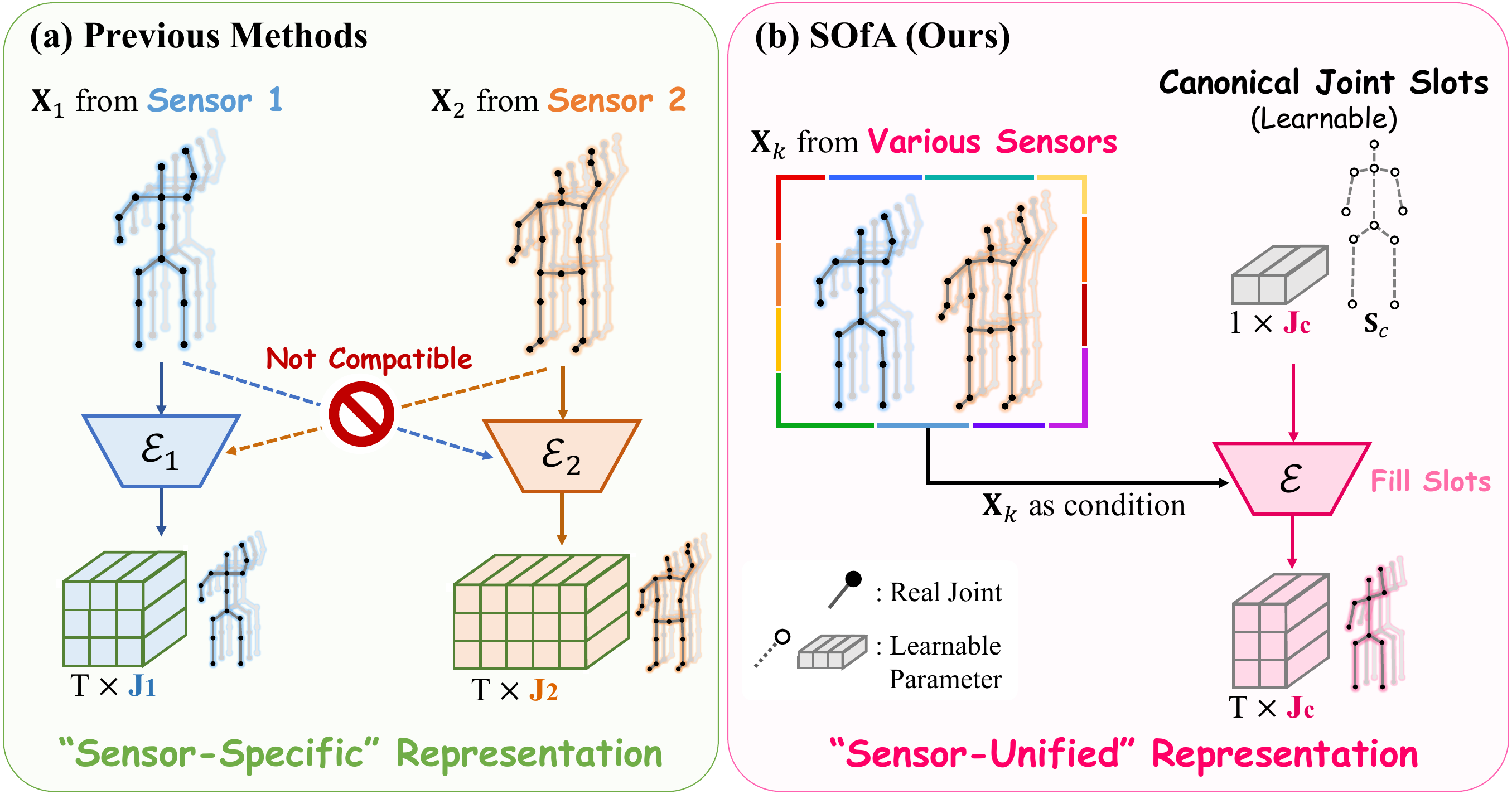}
  \vspace{-0.1cm}
  \caption{\textbf{Motivation of SOfA.} (a) Previous methods require isolated encoders for varying sensor topologies, yielding incompatible \textit{sensor-specific} representation. (b) SOfA (Ours) introduces fixed-size set of learnable Canonical Joint Slots, $\mathbf{S}_c$. By treating arbitrary sensor sequences as conditions, a single shared encoder dynamically fills these slots to extract a standardized, \textit{sensor-unified} representation, regardless of original joint counts.}
  \label{fig:motiv}
  \vspace{-0.3cm}
\end{wrapfigure}
To overcome the limitations of supervised learning, recent works in skeleton analysis have rapidly shifted toward Self-Supervised Learning (SSL) \citep{DINOv2, CL, MAE} to build foundation models capable of extracting universal motion representations from abundant unlabeled data. Fig.~\ref{fig:motiv} illustrates a conceptual comparison of skeleton representation learning paradigms. As shown in Fig.~\ref{fig:motiv}-(a), current skeleton SSL models fall short of being \textit{true} foundation models. While existing approaches \citep{MAMP, ActCLR, HSP, SJEPA, GFP} are individually trained and optimized for specific dataset structures, they claim foundation-level capabilities solely because a single encoder can be fine-tuned for various downstream tasks, even though the resulting representations remain \textit{strictly sensor- and dataset-specific}. Real-world skeleton sensors (e.g., Kinect \citep{Sensor}, LiDAR, and wearable sensors) exhibit vast diversity, yielding incompatible skeletal configurations and inconsistent anatomical definitions. Because current architectures are tightly coupled to a fixed number of joints and rigid absolute positional embeddings, they suffer from severe fragmentation; they are structurally incapable of processing out-of-distribution sensor data without being entirely redesigned and trained from scratch.

This fragmentation distinctly contrasts with the evolution of foundation models in other domains. In remote sensing \citep{SkySense, Copernicus}, for instance, foundation models successfully integrate heterogeneous modalities and varying resolutions from different satellites into a single, unified geo-spatial representation. The skeleton domain must similarly evolve toward a generalist architecture capable of encompassing diverse sensor configurations. At its core, a human action is defined by its underlying motion rather than the tool used to measure it; a ``jump'' remains the same movement whether it is recorded by 15 or 25 joints. Table~\ref{tab:comparison} provides a detailed comparison of properties across different skeleton representation learning frameworks. As shown, while a recent attempt, HSP \citep{HSP}, seeks to accommodate diverse skeleton formats, it remains limited to paired intra-scene multi-modal fusion rather than sensor-unified representation learning across independent sensors.

To address these fundamental limitations, we propose \textbf{SOfA} (\textbf{S}keleton \textbf{O}ne \textbf{f}or \textbf{A}ll), the \textit{first} generalist foundation model that achieves a \textit{sensor-unified representation} across \textit{arbitrary} configurations. To ensure universal adaptability, SOfA introduces a fixed-size set of learnable \textit{Canonical Joint Slots}. Conceptually inspired by the refinement process of diffusion models---where noise is iteratively distilled into a target output guided by conditioning signals---these slots act as a universal template that is dynamically populated by diverse, real-world skeleton sequences. Through an attention mechanism, SOfA distills disparate kinematic signals from various sensors into these standardized slots, ultimately yielding a unified representation that remains consistent regardless of the input's original topology as shown in Fig.~\ref{fig:motiv}-(b).

A remaining challenge is the misalignment of joint indices across diverse sensors. Traditional absolute positional embeddings assign rigid numerical indices that are essentially unreasonable for multi-sensor settings, as the same indices often refer to different anatomical joints across datasets, making it difficult for the model to learn a consistent structural representation. To resolve this, we leverage a pre-trained text encoder to introduce \textit{Semantic Joint Embedding}, which utilizes explicit textual names (e.g., ``Wrist'') from sensor metadata. This semantic-driven positional embedding enables the model to intrinsically capture anatomical correspondences, mitigating the impact of mismatched joint indexing between sensors. To facilitate cross-sensor research, we unify ten 3D skeleton datasets into a standardized benchmark. Our contributions are summarized as follows:

\begin{itemize}
    \item We propose \textbf{SOfA}, the \textit{sensor-unified foundation model} for skeleton representation learning capable of processing heterogeneous data across varying sensor types, joint counts, and indexing protocols within a single network.
    \item We introduce a novel paradigm to tackle topological heterogeneity via attention-based \textit{Canonical Joint Slots}, which are dynamically populated using arbitrary sensor inputs as conditioning signals.
    \item To align joint indices between heterogeneous sensors, we present \textit{Semantic Joint Embedding}, replacing rigid absolute positional embeddings with semantic features derived from textual labels via a pre-trained text encoder.
    \item Extensive experiments demonstrate that SOfA, functioning as a single universal encoder, achieves SOTA performance across diverse datasets and downstream tasks, yielding results that are highly comparable to or even outperform those of sensor-specific models.
\end{itemize}

%% file: sec/2_related.tex
\section{Related Works}
\label{sec:related}

\subsection{SSL for Skeleton Representations}
Recent advancements in self-supervised learning \citep{DINO, DINOv2, Register, CL, MAE} for skeleton data have evolved to extract rich motion features through strategies like Contrastive Learning (CL) \citep{CL} and Masked auto-encoder (MAE) \citep{MAE}. CL-based approaches focus on capturing instance-level discrimination, utilizing methods ranging from augmented view agreement \citep{CrosSCLR, AimCLR}) to advanced semantic mining \citep{ActCLR, HaLP}. Alternatively, MAE-based methods \citep{MAMP, SkeletonMAE, SJEPA, GFP, AMR} encourage models to capture local details by reconstructing raw coordinates \citep{SkeletonMAE, MAMP} or latent features \citep{SJEPA, GFP, AMR}. While these methodologies have driven significant progress, they share a critical structural bottleneck: they depend on fixed joint topologies and absolute positional embeddings. Consequently, they become structurally incompatible when an entirely new sensor with a different joint count is introduced, making them unable to serve as sensor-unified foundation models.

To tackle this data heterogeneity, a recent approach, HSP \citep{HSP}, attempts to integrate diverse skeleton formats by fusing 2D and 3D sequences. However, as highlighted in Table~\ref{tab:comparison}, HSP \citep{HSP} focuses on intra-scene multi-modal fusion strategy that relies on paired data. To align differing joint coordinates, it uses naive skeletal interpolation, which distorts precise anatomical geometry and restricts its scalability to independent real-world sensors. In contrast, our proposed SOfA seamlessly aligns topological heterogeneity without geometric distortion. With Canonical Joint Slots and Semantic Joint Embedding, SOfA establishes the first sensor-unified foundation model capable of universal generalization across completely independent sensors. To bridge this gap, SOfA establishes a sensor-unified foundation model that captures the intrinsic kinematic essence of human motion across diverse sensor configurations.

\subsection{Foundation Models in Other Fields}
The innovative success of foundation models in natural language processing (NLP) and 2D vision establishes a powerful example for unifying fragmented datasets through large-scale representation learning. In NLP, Large Language Models (LLMs) \citep{LLM} achieve exceptional flexibility by mapping highly diverse texts---regardless of languages, lengths, or structures---into a standardized sequence of sub-word tokens. Similarly, vision foundation models, such as DINO \citep{DINO, DINOv2} and SAM \citep{SAM}, achieve powerful zero-shot generalization by decoupling their architectures from fixed input resolutions. By treating images as flexible sequences of patch tokens rather than rigid pixel grids, they can seamlessly process inputs of arbitrary sizes and formats. In the 3D point cloud \citep{PointMAE, PointBERT} and remote sensing \citep{SkySense, Copernicus} domains, recent frameworks handle highly unconstrained and heterogeneous data by projecting them into a shared latent space.

Parallel to these advancements, human motion is fundamentally invariant across sensor modalities; the underlying dynamics of an action remain constant despite variations in the capture devices. To the best of our knowledge, none of the existing frameworks can effectively reconcile the structural and semantic differences arising from multi-sensor skeleton data within a single architecture.

%% file: sec/3_method.tex
\section{Skeleton One for All (SOfA)}
\label{sec:method}

\subsection{SOfA Framework}

\begin{figure*}[t]
  \centering
  \includegraphics[width=0.85\textwidth]{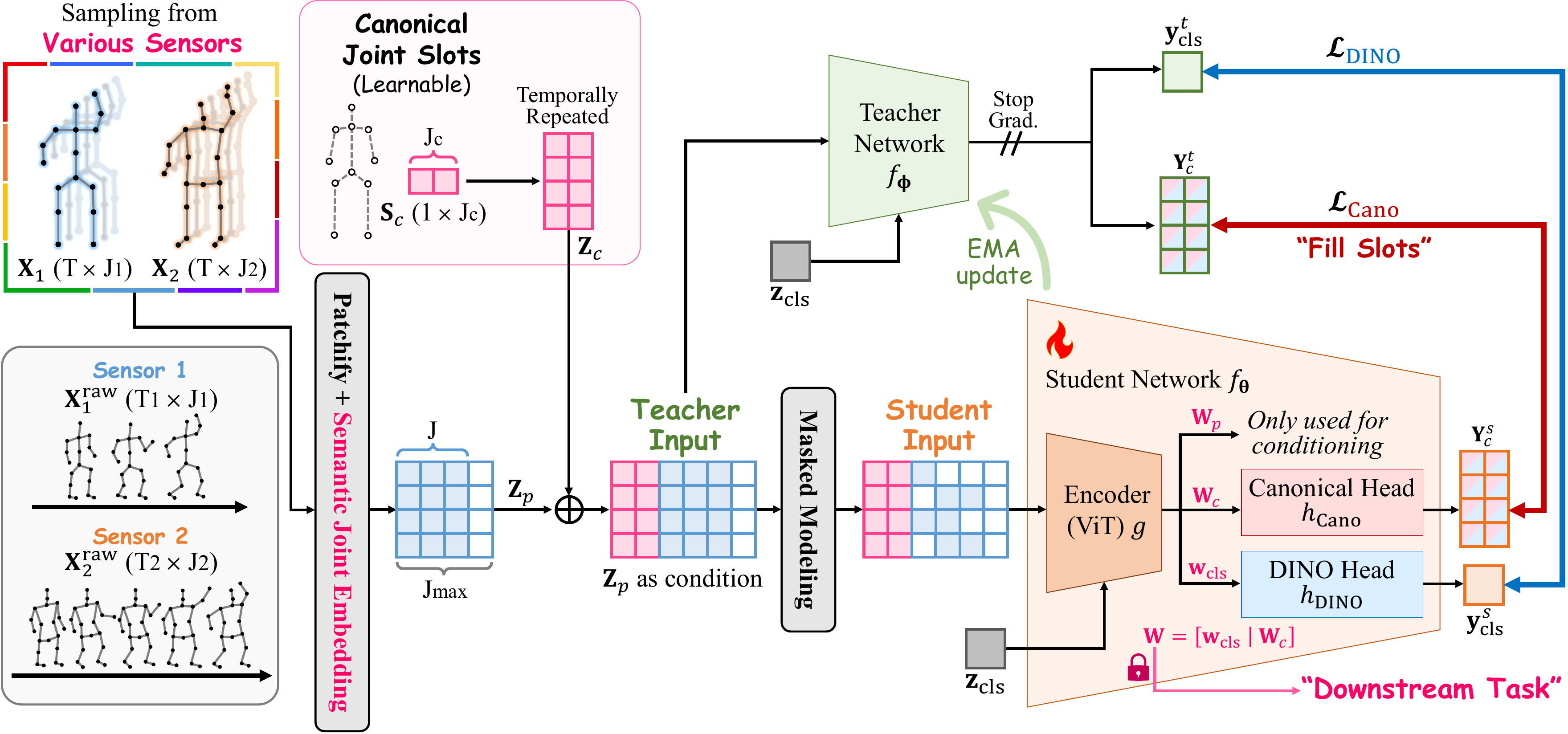}
  \caption{\textbf{Overview of SOfA.} Our framework establishes a sensor-unified representation within a teacher--student architecture, where both networks dynamically fill a fixed-size set of Canonical Joint Slots. Processing a masked view, the student encoder minimizes the canonical reconstruction error ($\mathcal{L}_{\text{Cano}}$) to match the slot representations of the unmasked teacher, alongside a global distillation loss ($\mathcal{L}_{\text{DINO}}$) to align high-level semantics. By treating varying sensor tokens $w_p$ as dynamic conditions, the model exclusively extracts the unified canonical representation $\mathbf{w}$ for downstream tasks.}
  \label{fig:framework}
\end{figure*}

\noindent\textbf{Overview.} Fig.~\ref{fig:framework} illustrates the overall architecture of \textbf{SOfA}, a sensor-unified framework built upon a teacher--student distillation paradigm \citep{Teacher}. In this framework, both the student network ($f_{\bm\theta}$) and teacher network ($f_{\bm\phi}$) share an identical encoder $g$ with lightweight (3-layer MLP) prediction heads $h_{\text{Cano}}$ and $h_{\text{DINO}}$. While the student network parameters $\bm\theta$ are updated via back-propagation, the teacher network parameters $\bm\phi$ evolves smoothly through an exponential moving average (EMA) of the student: $\bm\phi \leftarrow \tau\bm\phi + (1-\tau)\bm\theta$, governed by momentum $\tau$. Furthermore, SOfA addresses the limitations of sensor-specific models by bridging two structural barriers in multi-sensor data: (i) the \textit{joint index misalignment} (sensor-specific joint indexing); and (ii) the \textit{joint count discrepancy} (varying numbers of joints). Joint index misalignment is resolved through Semantic Joint Embedding (Sec.~\ref{sec:sje}), which aligns joints semantically across sensors. Joint count discrepancy is handled by a fixed-size set of Canonical Joint Slots (Sec.~\ref{sec:cjs}), which treats arbitrary sensor input as a dynamic condition to extract a unified canonical representation. This serves as the final, universally compatible output, regardless of the original sensor topology.

\noindent\textbf{Unified dataset formulation.} To ensure geometric consistency across heterogeneous sources, we first align all datasets into a standardized 3D coordinate system. For each sample, the ``hip'' joint of the initial frame---a universally shared anatomical landmark---is designated as the origin. In addition, the vertical body axis (head-to-foot) is aligned with the $y$-axis, and all coordinates are normalized to meters to maintain physical consistency. Formally, for an input sequence $\mathbf{X}$, we define the multi-sensor corpus as $\mathcal{D} = \{(\mathbf{X}_i^{\mathrm{raw}}, s_i)\}_{i=1}^N$, where $\mathbf{X}_i^{\text{raw}} \in \mathbb{R}^{T_i \times J_i \times 3}$ is $i$-th input skeleton sequence with temporal length $T_i$ and $J_i$ sensor-specific 3D joints. Each sensor identifier $s_i$ is associated with a metadata dictionary specifying the mapping between semantic joint names and their sensor-specific numerical indices. To enable batch-wise computation, we randomly sample $T$ frames and zero-pad the sequences along the joint dimension to $J_{\text{max}}$. Here, $J_{\text{max}}$ denotes the maximum number of joints across the entire corpus. This results in a standardized tensor $\mathbf{X} \in \mathbb{R}^{T \times J_{\text{max}} \times 3}$, providing a unified input format for our architecture.

\noindent\textbf{Architecture of the encoder.} Our framework is built upon a Vision Transformer (ViT) \citep{ViT} encoder $g$, which explicitly casts the raw skeleton sequence $\mathbf{X}$ as a \textit{dynamic condition input}. First, $\mathbf{X}$ is partitioned into a sequence of flattened 2D patches $\mathbf{X}_p \in \mathbb{R}^{N \times D_p}$, where $N = (T/P_T) \times (J_{\text{max}}/P_J)$ is the total number of patches with skeletal-temporal resolution $(P_T, P_J)$, and $D_p = P_T \cdot P_J \cdot 3$. To map these diverse sensor inputs into a shared space, each patch undergoes a linear projection $\mathbf{E} \in \mathbb{R}^{D_p \times D}$. Crucially, instead of absolute positional embedding, we inject our Semantic Joint Embedding (\texttt{SJE}, detailed in Sec.~\ref{sec:sje}) to assign explicit anatomical identities to condition tokens $\mathbf{Z}_p \in \mathbb{R}^{N \times D}$ as:
\begin{equation}
\label{eq:sje}
\mathbf{Z}_p =  \left[\mathbf{x}_p^{1}\mathbf{E}\mid\mathbf{x}_p^{2}\mathbf{E}\mid \cdots \mid \mathbf{x}_p^{N} \mathbf{E}\right] + \texttt{SJE}.
\end{equation}
Concurrently, we introduce a set of learnable Canonical Joint Slots (\texttt{CJS}, detailed in Sec.~\ref{sec:cjs}), denoted as $\mathbf{S}_c \in \mathbb{R}^{1 \times J_c \times D}$. To ensure temporal alignment of the condition input $\mathbf{Z}_p$, $\mathbf{S}_c$ is replicated $T/P_T$ times along the time axis, forming the canonical token $\mathbf{Z}_c \in \mathbb{R}^{N_c \times D}$ with $N_c = (T/P_T) \times J_c$. A learnable class token $\mathbf{z}_{\texttt{cls}}$ and canonical tokens $\mathbf{Z}_c$ are then prepended to the patch tokens $\mathbf{Z}_p$ to initialize the input stream as $\mathbf{Z}_0 = \left[\mathbf{z}_{\texttt{cls}} \mid \mathbf{Z}_c \mid \mathbf{Z}_p \right]$. To capture sequence-level dependencies, 1D temporal Rotary Positional Embeddings (RoPE) \citep{RoPE} are applied directly within the attention mechanism. The forward propagation across $L$ layers is formulated as:
\begin{equation}
\label{eq:attention}
\begin{aligned}
    &\mathbf{Z}_0 = \left[\mathbf{z}_{\texttt{cls}}\mid \mathbf{Z}_c\mid \mathbf{Z}_p\right], &&\quad \textit{Input initialization}\\
    &\mathbf{Z}_l^{\prime} = \mathsf{MSA}\left(\mathsf{LN}\left(\mathbf{Z}_{l-1}\right)\right) + \mathbf{Z}_{l-1}, &&\quad \textit{Attention w/ temporal RoPE} \\
    &\mathbf{Z}_l = \mathsf{MLP}\!\left(\mathsf{LN}\left(\mathbf{Z}_{l}^{\prime}\right)\right) + \mathbf{Z}_{l}^{\prime}\\
    &\left[\mathbf{w}_{\texttt{cls}}\mid \mathbf{W}_c \mid \mathbf{W}_p\right]=\mathsf{LN}\left(\mathbf{Z}_{L}\right), &&\quad \textit{Encoder output}\\
    &\mathbf{W} = \left[\mathbf{w}_{\texttt{cls}}\mid \mathbf{W}_c \right] &&\quad \textit{Final skeleton representation}
\end{aligned}
\end{equation}
where $\mathsf{MSA}$, $\mathsf{LN}$, and $\mathsf{MLP}$ denote Multi-Head Self-Attention, Layer Normalization, and the feed-forward network, respectively. After encoding, the sensor-specific tokens $\mathbf{W}_p$ are discarded, retaining only the unified canonical representation $\mathbf{W}$ for downstream tasks.

\noindent \textbf{Prediction heads.} The model employs two task-specific prediction heads, the canonical head $h_{\text{Cano}}$ and the contrastive head $h_{\text{DINO}}$ (depicted in Fig.~\ref{fig:framework}), to process the encoder outputs. $h_{\text{Cano}}$ maps the sensor-unified local representations $\mathbf{W}_c$ to target dimensions, yielding $\mathbf{Y}_c = h_{\text{Cano}}(\mathbf{W}_c)$. $h_{\text{DINO}}$ projects the global class token $\mathbf{w}_{\texttt{cls}}$ for instance discrimination, producing $\mathbf{y}_{\texttt{cls}} = h_{\text{DINO}}(\mathbf{w}_{\texttt{cls}})$. During training, the teacher network $f_{\bm\phi}$ processes the full, unmasked sequence to generate stable target representations: $\left[\mathbf{y}_{\texttt{cls}}^{t}\mid \mathbf{Y}_{c}^{t}\right] = f_{\bm\phi}\left(\left[\mathbf{z}_{\texttt{cls}}\mid \mathbf{Z}_{c} \mid \mathbf{Z}_{p}\right]\right)$. Conversely, the student network $f_{\bm\theta}$ receives a corrupted view. We apply a binary mask $\mathcal{M}_p \in \{0, 1\}^N$ to the patch tokens, defining the masked input $\mathbf{Z}_{p}^{\texttt{mask}}$ as:
\begin{equation}
    \mathbf{Z}_{p}^{\texttt{mask}} = \mathcal{M}_p \odot \mathbf{Z}_{p} + (1-\mathcal{M}_p) \odot \mathbf{e}_{\texttt{[mask]}}^b,
    \label{eq:mask}
\end{equation}
where $\mathbf{e}_{\texttt{[mask]}} \in \mathbb{R}^{D}$ is a learnable mask token, and $\mathbf{e}_{\texttt{[mask]}}^b \in \mathbb{R}^{N \times D}$ represents its broadcasted version to the masked positions. $f_{\bm\theta}$ then predicts features based on this corrupted context as: $\left[\mathbf{y}_{\texttt{cls}}^{s}\mid \mathbf{Y}_{c}^{s}\right] = f_{\bm\theta}\left(\left[\mathbf{z}_{\texttt{cls}}\mid \mathbf{Z}_{c} \mid \mathbf{Z}_{p}^{\texttt{mask}}\right]\right)$.

\noindent \textbf{Optimization.} We optimize SOfA through a dual-objective framework that enforces both global context and local structural alignment. For global context, we adopt a distillation loss inspired by DINO \citep{DINOv2}, aligning the instance-level representations between $f_{\bm\phi}$ and $f_{\bm\theta}$:
\begin{equation}
    \mathcal{L}_{\text{DINO}} = - \sum \mathbf{p}_{\texttt{cls}}^t \log \mathbf{p}_{\texttt{cls}}^s,
\end{equation}
where $\mathbf{p}_{\texttt{cls}}^t$ and $\mathbf{p}_{\texttt{cls}}^s$ are the softmax probability distributions derived from the class outputs $\mathbf{y}_{\texttt{cls}}^{t}$ and $\mathbf{y}_{\texttt{cls}}^{s}$. For the local objective, we adopt the Canonical Slot Reconstruction loss, $\mathcal{L}_{\text{Cano}}$. Crucially, since $f_{\bm\theta}$ processes only a corrupted subset of the sensor input, it must implicitly learn to \textit{fill} the fixed-size $\mathbf{Z}_c$ by inferring the missing kinematics from the visible condition patches. We formulate this slot-filling mechanism as an objective between $f_{\bm\theta}$'s predicted canonical representations and $f_{\bm\phi}$'s stable targets:
\begin{equation}
    \mathcal{L}_{\text{Cano}} = - \sum \mathbf{p}_{c}^t \log \mathbf{p}_{c}^s,
\end{equation}
where $\mathbf{p}_{c}^t$ and $\mathbf{p}_{c}^s$ denote the softmax distributions of the teacher and student canonical outputs, $\mathbf{y}_{c}^{t}$ and $\mathbf{y}_{c}^{s}$ respectively. The overall objective $\mathcal{L}_{\text{total}}$ integrates the canonical reconstruction and the global semantic alignment:
\begin{equation}
    \mathcal{L}_{\text{total}} = \mathcal{L}_{\text{Cano}} + \lambda\mathcal{L}_{\text{DINO}},
\end{equation}
where $\lambda$ is a hyperparameter balancing the two components. We empirically set $\lambda=1.0$ for all experiments.

\begin{figure*}[t]
  \centering
  \includegraphics[width=0.85\textwidth]{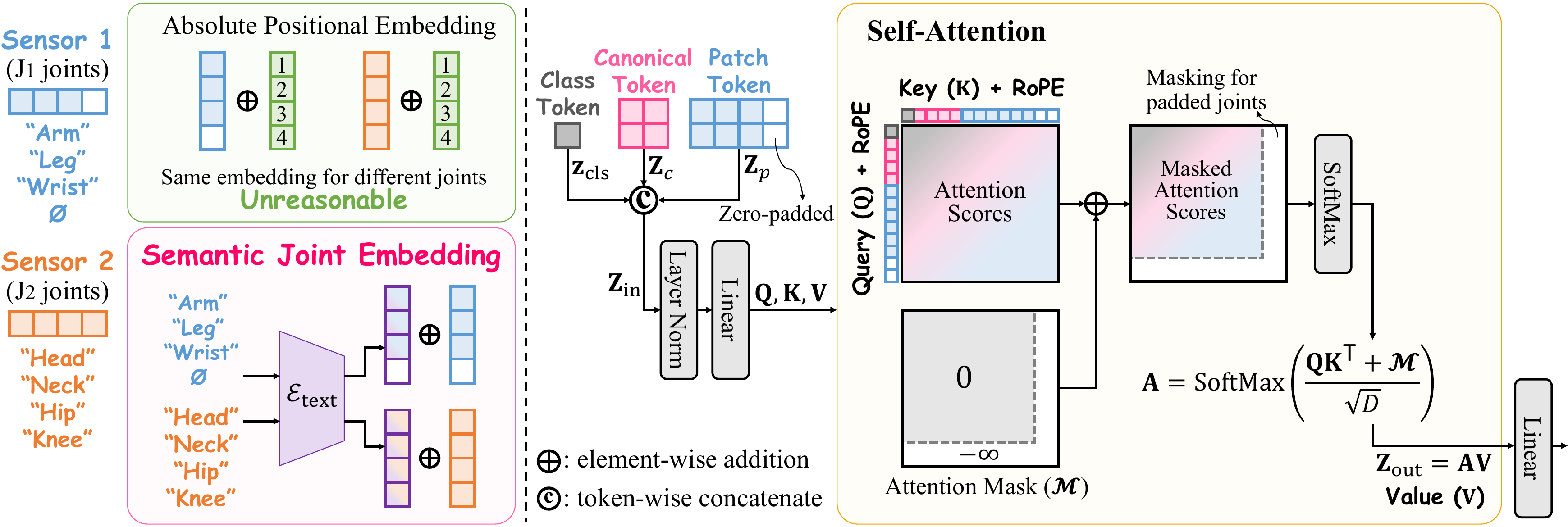}
  \caption{\textbf{Semantic Joint Alignment and Canonical Joint Slot aggregation.} (\textit{Left}) to resolve the joint index misalignment, Semantic Joint Embedding (\texttt{SJE}) utilizes a pretrained text encoder to map heterogeneous joint names into a shared latent space, ensuring anatomically consistent representations regardless of the sensor-specific numerical indices. (\textit{Right}) the attention mechanism jointly propagates the concatenated class token, canonical tokens, and condition patch tokens. This allows the model to effectively fill the canonical slots by aggregating information from the visible joints, while an attention mask is applied to exclude the influence of zero-padded tokens.}
  \label{fig:module}
\end{figure*}

\subsection{Semantic Joint Embedding (SJE)}
\label{sec:sje}
Fig.~\ref{fig:module} illustrates detailed mechanism of SOfA. We introduce \texttt{SJE} to mitigate the \textit{joint index misalignment} caused by sensor-specific topologies as shown in Fig.~\ref{fig:module}. Standard absolute positional embedding, which rely on arbitrary numerical indices, are fundamentally unreasonable for multi-sensor setups. In such systems, the same index often refers to different anatomical joints across datasets---or even to empty, zero-padded joints---preventing the model from learning a consistent structural representation. To tackle this, we utilize the explicit textual names provided in the metadata of each sensor. For each joint $j \in \{1, \dots, J_{\text{max}}\}$, we construct a descriptive prompt: $\mathbf{d}_j=\textit{``A human skeleton joint of the }\{n_j\}\textit{''}$, where $n_j$ denotes the $j$-th joint name (e.g., ``Left Wrist'', ``Head''). These prompts are processed through a pre-trained text encoder $\mathcal{E}_{\text{text}}$ to extract high-dimensional semantic vectors. Formally, the \texttt{SJE} for the $j$-th joint is defined as:
\begin{equation}
\texttt{SJE}_j =
\begin{cases}
\mathcal{E}_{\text{text}}\left(\mathbf{d}_j\right) &\quad\text{if } j \leq J_i \\
\mathbf{0} &\quad\text{if } j > J_i \text{ (zero-padded)}
\end{cases}
\end{equation}
where $J_i$ is the number of active joints for sensor $s_i$. We inject \texttt{SJE} into the input $\mathbf{Z}_p$ as described in Eq.~\ref{eq:sje}. This approach ensures that anatomically identical joints share a consistent latent identity across all datasets. Furthermore, using the pre-trained text encoder $\mathcal{E}_{\text{text}}$, the model inherits semantic relationships between joints. For instance, joints sharing the ``Left'' descriptor are mapped to a similar feature space, helping the network recognize shared functional roles within the human body.

\subsection{Canonical Joint Slots (CJS) and Self-Attention Mechanism}
\label{sec:cjs}

We address joint count discrepancy by introducing Canonical Joint Slots (CJS), a fixed-size set of learnable tokens $\mathbf{S}_c \in \mathbb{R}^{1 \times J_c \times D}$ that act as a universal vessel for all skeleton data. Inspired by multi-modal Diffusion Transformer (MM-DiT) \citep{DiT, MMDiT}, which simultaneously processes conditional signals and target features, we propagate the canonical tokens alongside the sensor-specific patch tokens. As shown in Fig.~\ref{fig:module}, the input stream is constructed by concatenating all tokens (Eq.~\ref{eq:attention}). We implement a \textit{slot-filling} mechanism via masked self-attention. Here, the canonical slots $\mathbf{Z}_c$ are dynamically populated by aggregating skeletal information from the condition tokens $\mathbf{Z}_p$. We employ an attention mask $\mathcal{M}$ to neutralize the influence of zero-padded joints within the $\mathbf{Z}_p$ dimension:
\begin{equation}
\mathbf{A} = \text{SoftMax}\left(\frac{\mathbf{QK}^\top + \mathcal{M}}{\sqrt{D}}\right),
\end{equation}
where $\mathcal{M} \in \{ 0, -\infty \}^{N_{\text{total}} \times N_{\text{total}}}$ is the masking matrix for a total sequence length of $N_{\text{total}} = 1 + N_c + N$, representing the $\mathbf{z}_{\texttt{cls}}$, $\mathbf{Z}_c$, and $\mathbf{Z}_p$, respectively. $\mathcal{M}$ assigns $-\infty$ to the attention scores corresponding to padded joint indices, effectively excluding invalid joints. This ensures that the final representation $\mathbf{W}$ is strictly computed from valid skeletal information, regardless of the input's original joint count. Beyond resolving joint-count discrepancy, the slot count $J_c$ offers a tunable output resolution; we analyze its effect and downstream flexibility in the \textit{Supplementary Material}.

\begin{table*}[t]
    \scriptsize
    \centering
    \caption{\textbf{Comparison with recent SSL methods on NTU-60, NTU-120, and PKU-MMD (25-joint).} Results are reported as Top-1 accuracy (\%) under the \textit{linear evaluation}. Unlike baseline methods that require independent training for each dataset, SOfA employs a single, unified representation across all benchmarks. \textbf{Bold} indicates the best result, and \underline{underline} indicates the second best. X-View$^{*}$ denotes evaluation on leakage-free test samples only (see \textit{Supplementary Material}).}
    \label{tab:main_result}
    \resizebox{0.85\textwidth}{!}{%
    \def\arraystretch{1.0}
    \setlength{\tabcolsep}{7pt}
    \begin{tabular}{llccccc}
    \toprule
    \multirow{2}{*}{\textbf{Method}} & \multirow{2}{*}{\textbf{Publication}} & \multicolumn{2}{c}{\textbf{NTU-60}} & \multicolumn{2}{c}{\textbf{NTU-120}} & \textbf{PKU-MMD} \\
    \cmidrule(lr){3-4} \cmidrule(lr){5-6} \cmidrule(lr){7-7}
     & & X-Sub & X-View & X-Sub & X-Set & X-Sub \\
    \midrule
    \multicolumn{7}{l}{\textbf{\textit{Sensor-Specific Representation}:}}\\
    GL-Transformer \citep{GL-Transformer} & ECCV'22 & 76.3 & 83.8 & 66.0 & 68.7 & -- \\
    CPM \citep{CPM} & ECCV'22 & 78.7 & 84.9 & 68.7 & 69.6 & 48.3 \\
    CMD \citep{CMD} & ECCV'22 & 79.8 & 86.9 & 70.3 & 71.5 & 43.0 \\
    AimCLR \citep{AimCLR} & AAAI'22 & 74.3 & 79.7 & 63.4 & 63.4 & -- \\
    HYSP \citep{HYSP} & ICLR'23 & 78.2 & 82.6 & 61.8 & 64.6 & -- \\
    HaLP \citep{HaLP} & CVPR'23 & 79.7 & 86.8 & 71.1 & 72.2 & 43.5 \\
    ActCLR \citep{ActCLR} & CVPR'23 & 80.9 & 86.7 & 69.0 & 70.5 & -- \\
    RVTCLR \citep{RVTCLR} & ICCV'23 & 74.7 & 79.1 & -- & -- & -- \\
    SkeletonMAE \citep{SkeletonMAE} & ICMEW'23 & 74.8 & 77.7 & 72.5 & 73.5 & 36.1 \\
    MAMP \citep{MAMP} & ICCV'23 & 84.9 & 89.1 & 78.6 & 79.1 & 53.8 \\
    PTSL \citep{PTSL} & AAAI'23 & 77.3 & 81.8 & 66.2 & 67.7 & 49.3 \\
    S-JEPA \citep{SJEPA} & ECCV'24 & 85.3 & 89.8 & 79.6 & 79.9 & 53.5 \\
    IGM \citep{IGM} & ECCV'24 & 86.2 & 91.2 & \underline{80.0} & 81.4 & -- \\
    MacDiff \citep{MacDiff} & ECCV'24 & 86.4 & 91.0 & 79.4 & 80.2 & -- \\
    HSP \citep{HSP} & CVPR'25 & 80.7 & 88.0 & 71.0 & 73.2 & 48.9 \\
    USDRL \citep{USDRL} & AAAI'25 & 85.2 & 91.7 & 76.6 & 78.1 & 54.4 \\
    GFP \citep{GFP} & ICCV'25 & 85.9 & 92.0 & 79.1 & 80.3 & 56.2 \\
    AMR \citep{AMR} & CVPR'26 & \textbf{87.4} & \underline{92.3} & \textbf{81.1} & \underline{81.9} & 60.3 \\
    \midrule
    \rowcolor{yellow!15}
    \multicolumn{7}{l}{\textbf{\textit{Sensor-Unified Representation}:}}\\
    \rowcolor{yellow!15}
    \textbf{SOfA (Ours)} & & 85.7 & $\text{92.1}^{*}$ & 79.2 & 80.4 & \underline{65.4} \\
    \rowcolor{yellow!15}
    \textbf{SOfA-L (Ours)} & & \underline{87.0} & $\textbf{92.6}^{*}$ & 79.5 & \textbf{82.0} & \textbf{67.1} \\
    \bottomrule
    \end{tabular}}
\end{table*}

%% file: sec/4_experiment.tex
\section{Experimental Results}
\label{sec:exp}

\subsection{Datasets}
\label{sec:dataset}
To pre-train and rigorously evaluate SOfA, we curate a large-scale corpus from various public benchmark datasets, encompassing a wide spectrum of sensor topologies and joint counts. A key contribution of our work is the standardization of these heterogeneous datasets into a unified coordinate system. Comprehensive profiles for each dataset and technical details regarding sensor-specific preprocessing are provided in the \textit{Supplementary Material}.

\noindent \textbf{Foundation model training.} For the large-scale pre-training of our foundation model, we utilize eight datasets with high joint counts (20 and 25 joints) to provide rich, fine-grained skeletal dynamics. This training corpus includes: (i) 25-joint datasets: NTU RGB+D 60 (NTU-60) \citep{NTU60}, NTU RGB+D 120 (NTU-120) \citep{NTU120}, PKU-MMD II (PKU-MMD) \citep{PKU}, ETRI-Activity3D (ETRI-Act) \citep{ETRI}, and ETRI-LivingLab \citep{ETRI}, which offer the most detailed skeletal topologies for complex action recognition; and (ii) 20-joint datasets: MSR-Action3D \citep{MSRAction3}, NW-UCLA \citep{NWUCLA}, and UT-Kinect \citep{UTKinect}, which serve to inject structural diversity and broaden sensor-specific variations during the pre-training phase.

\noindent \textbf{Unseen arbitrary sensor evaluation.} To validate SOfA's robustness against unseen sensor configurations, we reserve two 15-joint datasets for unseen sensor evaluation: Florence \citep{Florence} and SBU-Kinect-Interaction (SBU-Inter) \citep{SBU}. By evaluating these disparate sensor topologies with a frozen encoder $g$, we explicitly demonstrate the model's generalizability and adaptability.

\noindent \textbf{Pre-training corpus and train--test separation.} To prevent train--test contamination, we enforce sequence-level separation between pre-training and evaluation. During pre-training, since NTU-120 contains all NTU-60 sequences, we use only NTU-120 as the NTU source and retain only the intersection of the NTU-120 training splits (X-Sub $\cap$ X-Set). Evaluation on NTU-60 X-View requires an additional safeguard: X-View partitions sequences by camera ID, whereas NTU-120 X-Set partitions them by setup ID. The official X-View test split therefore cannot be used directly, as it contains some sequences present in the pre-training corpus. We exclude all such sequences before evaluation. Table~\ref{tab:main_result} reports accuracy on the remaining sequence-disjoint subset, denoted X-View$^{*}$. More details are in the \textit{Supplementary Material}.

\subsection{Experiment Details}
\label{sec:exp_details}
To ensure a fair comparison with previous sensor-specific methods, we employ an 8-layer (SOfA) and a 12-layer (SOfA-L) ViT \citep{ViT} as our backbone encoder $g$, with complexity comparable to that of previous methods. The encoder $g$ is configured with a hidden dimension of $D=256$ with $8$ attention heads for SOfA, and $D=384$ with $12$ attention heads for SOfA-L. Input skeleton sequences are fixed to a temporal length of $T=64$ frames. For patch embedding, we use $(P_T, P_J)=(8,1)$. We initialize $\mathcal{E}_{\text{text}}$ with a pretrained T5 text encoder \citep{T5}. Setting the maximum sensor size $J_{\text{max}}=25$, we fix the number of \texttt{CJS} to $J_c=5$. SOfA is pre-trained for 120 epochs within the teacher--student distillation framework \citep{Teacher}. The $f_{\bm\phi}$'s EMA momentum $\tau$ is gradually updated from 0.994 to 1. Optimization is performed using AdamW \citep{AdamW} with a base learning rate of $2.0 \times 10^{-4}$, incorporating a 20-epoch linear warmup followed by a cosine decay \citep{CosineAnneal} down to $1.0 \times 10^{-6}$. We train the model utilizing a total batch size of 1,536 across eight NVIDIA RTX A6000 GPUs.

\subsection{Evaluation on Heterogeneous Benchmarks}
\begin{wraptable}{r}{0.70\textwidth}
    \centering
    \vspace{-0.4cm}
    \scriptsize
    \caption{\textbf{Linear evaluation on seen benchmarks with varying joint
    configurations.} (a)~ETRI-Act (25-joint) and (b)~NW-UCLA (20-joint)
    are included in the pre-training domain. Results are Top-1 accuracy (\%)
    under the \textit{linear evaluation} protocol.}
    \vspace{-0.2cm}
    \label{tab:linear_seen}
    \begin{minipage}[t]{0.49\linewidth}
    \centering
    \textbf{\footnotesize (a) ETRI-Act (25-joint)}\\[2pt]
    \resizebox{\columnwidth}{!}{%
    \def\arraystretch{1.0}
    \setlength{\tabcolsep}{2pt}
    \begin{tabular}{lc}
        \toprule
        \textbf{Methods} & \textbf{ETRI-Act} \\
        \midrule
        \multicolumn{2}{l}{\textbf{\textit{Fully-supervised}:}}\\
        IndRNN \citep{IndRNN} & 73.9 \\
        Beyond Joint \citep{BeyondJoint} & 79.1 \\
        SK-CNN \citep{SKCNN} & 83.6 \\
        ST-GCN \citep{STGCN} & 86.8 \\
        Ensem-NN \citep{EnsemNN} & 83.0 \\
        MANs \citep{MANs} & 82.4 \\
        HCN \citep{HCN} & 88.0 \\
        FSA-CNN \citep{ETRI} & 90.6 \\
        \midrule
        \rowcolor{yellow!15}
        \multicolumn{2}{l}{\textbf{\textit{Un-supervised} (SSL):}}\\
        \rowcolor{yellow!15}
        \textbf{SOfA (Ours)} & 87.9 \\
        \bottomrule
    \end{tabular}}
    \end{minipage}
    \hfill
    \begin{minipage}[t]{0.49\linewidth}
    \centering
    \textbf{\footnotesize (b) NW-UCLA (20-joint)}\\[2pt]
    \resizebox{\columnwidth}{!}{%
    \def\arraystretch{1.0}
    \setlength{\tabcolsep}{2pt}
    \begin{tabular}{lc}
        \toprule
        \textbf{Methods} & \textbf{NW-UCLA} \\
        \midrule
        \multicolumn{2}{l}{\textbf{\textit{Sensor-Specific}:}}\\
        LongT-GAN \citep{LongTGAN} & 74.3 \\
        P\&C \citep{PandC} & 84.9 \\
        MCAE-MP \citep{MCAE} & 84.9 \\
        SeBiReNet \citep{SeBiReNet} & 80.3 \\
        Colorization \citep{Colorization} & 91.1 \\
        GL-Transformer \citep{GL-Transformer} & 90.4 \\
        Masked-Color \citep{MaskedColor} & 92.0 \\
        \midrule
        \rowcolor{yellow!15}
        \multicolumn{2}{l}{\textbf{\textit{Sensor-Unified}:}}\\
        \rowcolor{yellow!15}
        \textbf{SOfA (Ours)} & 92.9 \\
        \bottomrule
    \end{tabular}}
    \end{minipage}
    \vspace{-0.3cm}
\end{wraptable}
Table~\ref{tab:main_result} and Table~\ref{tab:linear_seen} summarize the linear evaluation results on both 25-joint and 20-joint benchmarks. As shown in Table~\ref{tab:main_result}, SOfA achieves competitive or superior performance on large-scale benchmarks like NTU-60 and NTU-120. To ensure a fair comparison, we report all results using the joint modality alone, without any multi-stream ensemble. Notably, existing skeleton-based SSL methods are inherently sensor-specific and even dataset-specific. They require training independent, isolated models for each unique protocol. In contrast, SOfA is a single, unified model trained across all datasets simultaneously. We provide extensive experimental results in the \textit{Supplementary Material}, including evaluations on semi-supervised learning, action retrieval, robustness across various sensor configurations, and a detailed computational-efficiency analysis showing that SOfA attains a 6.59$\times$ reduction in inference GFLOPs (4.30 vs.\ 28.32) over mainstream MAE-based SSL methods through aggressive temporal compression.

\noindent\textbf{Knowledge transfer in data-scarce scenarios.} On PKU-MMD, SOfA-L surpasses the previous best specialist (AMR, 60.3\%) by +6.8\%-point, reaching 67.1\%. While data-rich benchmarks like NTU-120 let specialists saturate by overfitting to massive sample sizes, the smaller PKU-MMD limits isolated models; SOfA bridges this gap by transferring robust skeletal priors from our diverse corpus, internalizing generalized human kinetics rather than memorizing sensor-specific patterns.

\noindent\textbf{Competitive strength against fully-supervised methods.} SOfA is also strong against fully-supervised models: as shown in Table~\ref{tab:linear_seen}(a), it reaches 87.9\% on ETRI-Act, within a close margin of high-performing supervised models---despite learning these features self-supervised, without any action labels during pre-training.

\noindent\textbf{Adaptability to various joint counts.} Table~\ref{tab:linear_seen}(b) shows SOfA's versatility across diverse skeletal topologies. On the 20-joint NW-UCLA dataset, SOfA achieves 92.9\% accuracy, outperforming all existing sensor-specific models. Crucially, this result is obtained using the exact same model weights employed for the 25-joint benchmarks in Table~\ref{tab:main_result} and Table~\ref{tab:linear_seen}(a). This success across various joint counts rigorously validates SOfA's capability as a unified foundation model that can seamlessly navigate heterogeneous sensor protocols without the need for any per-dataset reconfiguration.

\subsection{Generalization to Unseen Skeletal Protocols}

\begin{wraptable}{r}{0.70\textwidth}
    \centering
    \vspace{-0.4cm}
    \scriptsize
    \caption{\textbf{Linear evaluation on unseen cross-domain benchmarks.}
    (a)~Florence (15-joint) and (b)~SBU-Inter (15-joint)
    are excluded from pre-training and evaluates sensor-unified representation generalization. 
    Results are Top-1 accuracy (\%) under the \textit{linear evaluation}.}
    \vspace{-0.2cm}
    \label{tab:linear_unseen}
    \begin{minipage}[t]{0.50\linewidth}
    \centering
    \textbf{\footnotesize (a) Florence (15-joint)}\\[2pt]
    \resizebox{\columnwidth}{!}{%
    \def\arraystretch{1.0}
    \setlength{\tabcolsep}{6pt}
    \begin{tabular}{lc}
        \toprule
        \textbf{Methods} & \textbf{Florence} \\
        \midrule
        \multicolumn{2}{l}{\textbf{Seen+\textit{Fully-supervised}:}}\\
        Seidenari \textit{et al.}~\citeyearpar{seidenari2013recognizing} & 82.0 \\
        Devanne \textit{et al.}~\citeyearpar{devanne20143} & 87.0 \\
        Vemulapalli \textit{et al.}~\citeyearpar{vemulapalli2014human} & 90.9 \\
        HarSkel \citep{luvizon2017learning} & 94.4 \\
        \midrule
        \multicolumn{2}{l}{\textbf{Unseen+\textit{Sensor-Specific}:}}\\
        SkeletonMAE \citep{SkeletonMAE} & 66.8 \\
        MAMP \citep{MAMP} & 73.7 \\
        \midrule
        \rowcolor{yellow!15}
        \multicolumn{2}{l}{\textbf{Unseen+\textit{Sensor-Unified}:}}\\
        \rowcolor{yellow!15}
        \textbf{SOfA (Ours)} & 91.7 \\
        \bottomrule
    \end{tabular}}
    \end{minipage}
    \hfill
    \begin{minipage}[t]{0.48\linewidth}
    \centering
    \textbf{\footnotesize (b) SBU-Inter (15-joint)}\\[2pt]
    \resizebox{\columnwidth}{!}{%
    \def\arraystretch{1.0}
    \setlength{\tabcolsep}{6pt}
    \begin{tabular}{lc}
        \toprule
        \textbf{Methods} & \textbf{SBU-Inter} \\
        \midrule
        \multicolumn{2}{l}{\textbf{Seen+\textit{Fully-supervised}:}}\\
        Co-LSTM \citep{CoLSTM} & 90.4 \\
        ST-LSTM \citep{STLSTM} & 93.3 \\
        VA-LSTM \citep{VALSTM} & 97.2 \\
        GCA \citep{GCA} & 94.9 \\
        LSTM-IRN \citep{LSTMIRN} & 98.2 \\
        IGFormer \citep{IGFormer} & 98.4 \\
        ISTA-Net \citep{ISTA} & 98.5 \\
        \midrule
        \multicolumn{2}{l}{\textbf{Unseen+\textit{Sensor-Specific}:}}\\
        SkeletonMAE \citep{SkeletonMAE} & 73.1 \\
        MAMP \citep{MAMP} & 80.2 \\
        \midrule
        \rowcolor{yellow!15}
        \multicolumn{2}{l}{\textbf{Unseen+\textit{Sensor-Unified}:}}\\
        \rowcolor{yellow!15}
        \textbf{SOfA (Ours)} & 97.0 \\
        \bottomrule
    \end{tabular}}
    \end{minipage}
    \vspace{-0.3cm}
\end{wraptable}
To evaluate the true robustness of SOfA as a foundation model, we test it on arbitrary out-of-distribution (OOD) sensors completely excluded from pre-training. Table~\ref{tab:linear_unseen}(a) and Table~\ref{tab:linear_unseen}(b) report results on the unseen Florence and SBU-Inter datasets. Without any fine-tuning, SOfA reaches 91.7\% on Florence and 97.0\% on SBU-Inter, whereas sensor-specific SSL methods such as SkeletonMAE and MAMP degrade sharply (e.g., 66.8\% and 73.7\% on Florence). Notably, SOfA matches fully-supervised models trained end-to-end with full label access, providing solid evidence that it transcends sensor-specific limitations by internalizing universal skeletal rules.

\subsection{Ablation Studies}

\noindent\textbf{Importance of canonical space via CJS.} Without \texttt{CJS}, the model must aggregate all information from arbitrary sensor protocols into a single class token $\mathbf{z}_{\texttt{cls}}$, creating an information bottleneck that cannot capture the complex spatio-temporal dynamics of diverse skeletal structures. As shown in Table~\ref{tab:ablation}, removing \texttt{CJS} sharply degrades performance across all benchmarks (average drop of 8.48\%), confirming that our canonical slots provide an essential structured, high-resolution workspace for universal motion representation.

\begin{wraptable}{r}{0.61\textwidth}
    \centering
    \vspace{-0.4cm}
    \scriptsize
    \caption{\textbf{Ablation of SOfA components.} We isolate the contributions of \texttt{CJS} and \texttt{SJE} across heterogeneous topologies.}
    \label{tab:ablation}
    \vspace{-0.2cm}
    \resizebox{0.58\textwidth}{!}{%
    \def\arraystretch{1.0}
    \setlength{\tabcolsep}{3pt}
    \begin{tabular}{ccccccc}
        \toprule
        \multicolumn{2}{c}{\textbf{Modules}} & \multicolumn{3}{c}{\textbf{25-Joint Datasets}} & \multicolumn{2}{c}{\textbf{20-Joint Datasets}} \\
        \cmidrule(lr){1-2} \cmidrule(lr){3-5} \cmidrule(lr){6-7}
        \texttt{CJS} & \texttt{SJE} & NTU-60 & NTU-120 & PKU-MMD & NW-UCLA & UT-Kinect \\
        \midrule
        \ding{55} & \ding{55} & 75.2 & 70.7 & 50.9 & 68.9 & 77.0\\
        \ding{51} & \ding{55} & 84.9 & 78.7 & 64.7 & 72.3 & 80.0 \\
        \ding{55} & \ding{51} & 76.4 & 71.4 & 52.3 & 86.7 & 91.0 \\
        \rowcolor{yellow!15}
        \ding{51} & \ding{51} & 85.7 & 79.2 & 65.4 & 92.9 & 97.0 \\
        \bottomrule
    \end{tabular}}
    \vspace{-0.3cm}
\end{wraptable}

\noindent\textbf{Geometric reasoning via SJE.} The role of \texttt{SJE} is particularly critical for heterogeneous topologies. Removing \texttt{SJE} causes a catastrophic drop on 20-joint datasets (e.g., NW-UCLA from 92.9\% to 72.3\%), while the impact on 25-joint datasets is moderate. This stems from the data distribution: roughly 90\% of our corpus is 25-joint Kinect-v2 data, so without \texttt{SJE} the network becomes biased toward the 25-joint indexing, and \texttt{SJE} acts as the semantic anchor bridging this index discrepancy.

%% file: sec/5_conclusion.tex
\section{Conclusion}

We present \textbf{SOfA}, the first sensor-unified foundation model for skeleton representation learning that bridges heterogeneous sensor types, joint counts, and indexing protocols. Shifting from dataset-specific modeling to a unified architecture, SOfA addresses topological heterogeneity through attention-based \textit{Canonical Joint Slots} and \textit{Semantic Joint Embeddings}, reaching state-of-the-art performance across diverse benchmarks with a single unified encoder.

%% file: sec/6_appendix.tex
\appendix

\section*{Appendix}
\noindent This \textit{Supplementary Material} provides additional technical details, dataset profiles, and experimental results that complement the main paper. First, Sec.~\ref{sec:add_result} provides further empirical validation, including network scalability, an extended synthetic topology, semi-supervised learning, action retrieval, and linear evaluation on 20-joint datasets, together with a computational-complexity comparison against state-of-the-art methods. Sec.~\ref{sec:add_ablation} presents additional ablation studies on the number of Canonical Joint Slots (\texttt{CJS}) and analyzes their interpretability. Sec.~\ref{sec:leakage} details our pre-training corpus construction and the leakage-free X-View$^{*}$ protocol. Sec.~\ref{sec:dataset_detail} offers dataset summaries, coordinate statistics, and preprocessing details for the ten heterogeneous datasets. Finally, Sec.~\ref{sec:implement} elaborates on implementation specifics.

\begin{table}[h]
    \scriptsize
    \centering
    \caption{Overview of the \textit{Supplementary Material}.}
    \label{tab:supple_overview}
    \resizebox{0.45\columnwidth}{!}{%
    \def\arraystretch{1.0}
    \setlength{\tabcolsep}{6pt}
    \begin{tabular}{cl}
        \toprule
        \rowcolor{yellow!15}
        \textbf{Section} & \textbf{Contents} \\
        \midrule
        Sec.~\ref{sec:add_result} & Additional results and discussion \\
        Sec.~\ref{sec:add_ablation} & More ablation studies on \texttt{CJS} \\
        Sec.~\ref{sec:leakage} & Pre-training corpus and data leakage \\
        Sec.~\ref{sec:dataset_detail} & Datasets and preprocessing \\
        Sec.~\ref{sec:implement} & Implementation details \\
        \bottomrule
    \end{tabular}}
\end{table}

\section{Additional Results and Discussion}
\label{sec:add_result}

\begin{table}[t]
    \scriptsize
    \centering
    \caption{Network scalability on the 25-joint benchmarks (NTU-60, NTU-120, PKU-MMD). Results are Top-1 accuracy (\%) under the \textit{linear evaluation} protocol using the joint (J) modality. \textbf{SOfA} is the base model, \textbf{SOfA} ($J{=}30$) is evaluated on an extended synthetic topology, and \textbf{SOfA-L} is the deeper and wider variant. \textbf{Bold} denotes the best result.}
    \label{tab:supple_result}
    \resizebox{0.7\columnwidth}{!}{%
    \def\arraystretch{1.0}
    \setlength{\tabcolsep}{4pt}
    \begin{tabular}{lcccccc}
    \toprule
    \multirow{2}{*}{\textbf{Method}} & \multirow{2}{*}{\textbf{Modality}} & \multicolumn{2}{c}{\textbf{NTU-60}} & \multicolumn{2}{c}{\textbf{NTU-120}} & \textbf{PKU-MMD} \\
    \cmidrule(lr){3-4} \cmidrule(lr){5-6} \cmidrule(lr){7-7}
     & & X-Sub & X-View & X-Sub & X-Set & X-Sub \\
    \midrule
    \rowcolor{yellow!15}
    \textbf{SOfA} & J & 85.7 & $\text{92.1}^{*}$ & 79.2 & 80.4 & 65.4 \\
    \rowcolor{yellow!15}
    \textbf{SOfA} ($J{=}30$) & J & 85.5 & $\text{91.9}^{*}$ & 79.0 & 80.1 & 65.6 \\
    \rowcolor{yellow!15}
    \textbf{SOfA-L} & J & \textbf{87.0} & $\text{\textbf{92.6}}^{*}$ & \textbf{79.5} & \textbf{82.0} & \textbf{67.1} \\
    \bottomrule
    \end{tabular}}
\end{table}

\begin{table}[t]
    \scriptsize
    \centering
    \caption{Computational complexity comparison on PKU-MMD. We report the number of tokens, training and inference GFLOPs, and Top-1 accuracy (X-Sub). SOfA achieves state-of-the-art accuracy while substantially reducing inference cost.}
    \label{tab:flops}
    \resizebox{0.7\columnwidth}{!}{%
    \def\arraystretch{1.0}
    \setlength{\tabcolsep}{5pt}
    \begin{tabular}{lcccc}
    \toprule
    \multirow{2}{*}{\textbf{Method}} & \textbf{Tokens} & \multicolumn{2}{c}{\textbf{GFLOPs}} & \textbf{PKU-MMD} \\
    \cmidrule(lr){2-2} \cmidrule(lr){3-4} \cmidrule(lr){5-5}
    & $T \times J$ & Train & Inference & X-Sub \\
    \midrule
    \multicolumn{5}{l}{\textbf{\textit{Sensor-Specific}:}}\\
    SkeletonMAE \citep{SkeletonMAE} & $30 \times 25$ & 19.67 & 28.32 & 36.1 \\
    MAMP \citep{MAMP} & $30 \times 25$ & 19.67 & 28.32 & 53.8 \\
    S-JEPA \citep{SJEPA} & $30 \times 25$ & 47.99 & 28.32 & 53.5 \\
    GFP \citep{GFP} & $30 \times 25$ & 4.18 & 28.32 & 56.2 \\
    \midrule
    \rowcolor{yellow!15}
    \multicolumn{5}{l}{\textbf{\textit{Sensor-Unified}:}}\\
    \rowcolor{yellow!15}
    \textbf{SOfA} & $8 \times (25{+}5)$ & 8.60 & \textbf{4.30} & 65.4 \\
    \rowcolor{yellow!15}
    \textbf{SOfA-L} & $8 \times (25{+}5)$ & 14.90 & 7.45 & \textbf{67.1} \\
    \bottomrule
    \end{tabular}}
\end{table}

\subsection{Network Scalability}
\label{sec:large}

Table~\ref{tab:supple_result} presents the effect of scaling the encoder $g$. To evaluate the scalability of SOfA, we enlarge the backbone from the \textit{base configuration} (\textbf{SOfA}: 8 Transformer layers, hidden dimension 256) to a \textit{deeper and wider variant} (\textbf{SOfA-L}: 12 Transformer layers, hidden dimension 384). While the base model is adopted for fair comparison with prior methods \citep{MAMP, GFP, SkeletonMAE, SJEPA}, our pre-training corpus is substantially larger and more diverse than any single benchmark, covering eight heterogeneous datasets and roughly 2.5$\times$ more samples than NTU-120 alone. This motivates investigating whether SOfA can further benefit from increased model capacity. As shown in Table~\ref{tab:supple_result}, scaling the encoder consistently improves performance across all benchmarks, with SOfA-L yielding clear gains on NTU (e.g., NTU-60 X-Sub 85.7\%~$\rightarrow$~87.0\%, NTU-120 X-Set 80.4\%~$\rightarrow$~82.0\%). Due to GPU memory constraints, we reduce the total training batch size from 1,536 to 1,024 for SOfA-L. These results indicate that SOfA scales favorably with capacity and that the unified corpus is large enough to support stronger backbones, suggesting that the modest NTU gains of the base model stem from a capacity bottleneck rather than a limitation of the framework.

\subsection{Extended Synthetic Topology}
\label{sec:synthetic}
Zero-padding to $J_{\text{max}}=25$ is strictly a batching convenience, not an architectural ceiling. SOfA processes unseen topologies with more than 25 joints without retraining, owing to (i) an attention backbone that treats joints as a variable-length sequence (unlike GCNs with fixed adjacency matrices), and (ii) \texttt{SJE}, which dynamically generates valid embeddings for novel joints on the fly from their textual descriptions. We empirically verify this on an extended synthetic topology ($J=30$), created by interpolating pairs of joints ($J_a$ and $J_b$) and feeding their textual descriptions (e.g., ``midpoint of $J_a$ and $J_b$'') to \texttt{SJE}. As reported in Table~\ref{tab:supple_result}, the $J{=}30$ variant attains performance comparable to the standard $J{=}25$ setting, demonstrating that the architecture is agnostic to the padded joint budget.

\subsection{Computational Complexity}
\label{sec:flops}
Table~\ref{tab:flops} compares the computational cost of SOfA against mainstream MAE-based SSL methods \citep{SkeletonMAE, MAMP, SJEPA, GFP}. Although SOfA introduces additional canonical tokens along the joint dimension, its aggressive temporal compression yields a substantially lower overall cost: SOfA requires only 4.30 GFLOPs at inference, a 6.59$\times$ reduction relative to the 28.32 GFLOPs of MAE-based baselines, while simultaneously improving Top-1 accuracy on PKU-MMD by a large margin (65.4\% vs.\ 56.2\% for the strongest baseline). Even the larger SOfA-L, at 7.45 GFLOPs, remains far cheaper than the baselines. For downstream deployment, the model can be further streamlined to use only the $8\times5$ canonical tokens, providing a scalable and lightweight backbone for real-time skeleton analysis.

\begin{table}[t]
    \centering
    \scriptsize
    \begin{minipage}[t]{0.48\columnwidth}
        \centering
        \caption{\textit{Semi-supervised} results on NTU-60 using only 1\% labeled data.}
        \label{tab:semi_supervised}
        \resizebox{\columnwidth}{!}{%
        \def\arraystretch{1.0}
        \setlength{\tabcolsep}{4pt}
        \begin{tabular}{lcc}
            \toprule
            \multirow{2}{*}{\textbf{Methods}} & \multicolumn{2}{c}{\textbf{NTU-60}} \\
            \cmidrule(lr){2-3}
             & X-Sub & X-View \\
            \midrule
            \multicolumn{3}{l}{\textbf{\textit{Sensor-Specific}:}}\\
            CPM \citep{CPM} & 56.7 & 57.5 \\
            CMD \citep{CMD} & 50.6 & 53.0 \\
            HaLP \citep{HaLP} & 46.6 & 48.7 \\
            HiCo \citep{HiCo} & 54.4 & 54.8 \\
            UmURL \citep{UmURL} & 58.1 & 58.3 \\
            SkeletonMAE \citep{SkeletonMAE} & 54.4 & 54.6 \\
            MAMP \citep{MAMP} & 66.0 & 68.7 \\
            S-JEPA \citep{SJEPA} & 67.5 & 69.1 \\
            USDRL \citep{USDRL} & 57.3 & 60.7 \\
            GFP \citep{GFP} & \textbf{71.8} & \underline{72.9} \\
            \midrule
            \rowcolor{yellow!15}
            \multicolumn{3}{l}{\textbf{\textit{Sensor-Unified}:}}\\
            \rowcolor{yellow!15}
            \textbf{SOfA} & \underline{71.6} & $\textbf{\text{74.2}}^{*}$ \\
            \bottomrule
        \end{tabular}}
    \end{minipage}
    \hfill
    \begin{minipage}[t]{0.48\columnwidth}
        \centering
        \caption{\textit{Action retrieval} results on NTU-60 (X-Sub, X-View).}
        \label{tab:action_retrieval}
        \resizebox{\columnwidth}{!}{%
        \def\arraystretch{1.0}
        \setlength{\tabcolsep}{4pt}
        \begin{tabular}{lcc}
            \toprule
            \multirow{2}{*}{\textbf{Methods}} & \multicolumn{2}{c}{\textbf{NTU-60}} \\
            \cmidrule(lr){2-3}
             & X-Sub & X-View \\
            \midrule
            \multicolumn{3}{l}{\textbf{\textit{Sensor-Specific}:}}\\
            LongT-GAN \citep{LongTGAN} & 39.1 & 48.1 \\
            P\&C \citep{PandC} & 50.7 & 76.3 \\
            ISC \citep{ISC} & 62.5 & 82.6 \\
            HaLP \citep{HaLP} & 65.8 & 83.6 \\
            HiCo \citep{HiCo} & 68.3 & 84.8 \\
            MAMP \citep{MAMP} & 62.0 & 70.0 \\
            GFP \citep{GFP} & \underline{70.9} & \textbf{87.1} \\
            \midrule
            \rowcolor{yellow!15}
            \multicolumn{3}{l}{\textbf{\textit{Sensor-Unified}:}}\\
            \rowcolor{yellow!15}
            \textbf{SOfA} & \textbf{72.1} & $\underline{\text{86.8}}^{*}$ \\
            \bottomrule
        \end{tabular}}
    \end{minipage}
\end{table}

\subsection{Semi-Supervised Learning}
Table~\ref{tab:semi_supervised} shows the semi-supervised comparison on NTU-60 \citep{NTU60}, using only 1\% of the training labels. SOfA achieves 71.6\% on X-Sub and 74.2\% on X-View, outperforming all previous sensor-specific methods on X-View and remaining highly competitive on X-Sub (only 0.2\%-point below GFP \citep{GFP}). This shows that a sensor-unified representation not only generalizes across heterogeneous topologies but also surpasses strong sensor-specific pre-training under extremely label-scarce conditions, indicating that our unified canonical representation captures rich, linearly separable motion semantics even with limited supervision.

\subsection{Action Retrieval}
Table~\ref{tab:action_retrieval} reports skeleton-based action retrieval on NTU-60 \citep{NTU60}. SOfA achieves 72.1\% on X-Sub and 86.8\% on X-View. On X-Sub it outperforms the previous best sensor-specific method, GFP \citep{GFP}, by 1.2\%-point (72.1\% vs.\ 70.9\%), establishing a new state of the art; on X-View it trails GFP by only 0.3\%-point. These results further demonstrate that a sensor-unified representation can match or exceed sensor-specific models in instance-level retrieval, indicating a well-structured latent space with strong discriminability.

\subsection{Linear Evaluation on 20-Joint Datasets}
Tables~\ref{tab:msr} and \ref{tab:utkinect} report linear evaluation on two 20-joint benchmarks, MSR-Action3D \citep{MSRAction3} and UT-Kinect \citep{UTKinect}. On MSR-Action3D, SOfA achieves 82.2\%, comparable to fully supervised methods and only 2.3\%-point below the best reported result (84.5\%). On UT-Kinect, SOfA attains 97.0\%, surpassing all compared fully supervised methods, including the previous best (96.5\%), by 0.5\%-point. These results further demonstrate that SOfA learns transferable, highly discriminative representations even on 20-joint protocols.

\begin{table}[t]
    \centering
    \scriptsize
    \begin{minipage}[t]{0.52\columnwidth}
    \centering
    \caption{Fully supervised comparison on MSR-Action3D (20-joint), \textit{linear eval}.}
    \label{tab:msr}
    \resizebox{\columnwidth}{!}{%
    \def\arraystretch{1.0}
    \setlength{\tabcolsep}{4pt}
    \begin{tabular}{lc}
        \toprule
        \textbf{Methods} & \textbf{MSR-Action3D} \\
        \midrule
        \multicolumn{2}{l}{\textbf{\textit{Fully-supervised}:}}\\
        HON4D \citep{HON4D} & 82.2 \\
        Rahmani \textit{et al.} \citep{Rahmani} & 82.7 \\
        Tran \textit{et al.} \citep{Tran} & 84.5 \\
        \midrule
        \rowcolor{yellow!15}
        \multicolumn{2}{l}{\textbf{\textit{Un-supervised}:}}\\
        \rowcolor{yellow!15}
        \textbf{SOfA} & 82.2 \\
        \bottomrule
    \end{tabular}}
    \end{minipage}
    \hfill
    \begin{minipage}[t]{0.44\columnwidth}
        \centering
        \caption{Fully supervised comparison on UT-Kinect (20-joint), \textit{linear eval}.}
        \label{tab:utkinect}
        \resizebox{\columnwidth}{!}{%
        \def\arraystretch{1.0}
        \setlength{\tabcolsep}{4pt}
        \begin{tabular}{lc}
            \toprule
            \textbf{Methods} & \textbf{UT-Kinect} \\
            \midrule
            \multicolumn{2}{l}{\textbf{\textit{Fully-supervised}:}}\\
            Xia \textit{et al.} \citep{Xia} & 90.9 \\
            Devanne \textit{et al.} \citep{devanne20143} & 91.5 \\
            Wang \textit{et al.} \citep{Wang} & 96.5 \\
            \midrule
            \rowcolor{yellow!15}
            \multicolumn{2}{l}{\textbf{\textit{Un-supervised}:}}\\
            \rowcolor{yellow!15}
            \textbf{SOfA} & 97.0 \\
            \bottomrule
        \end{tabular}}
    \end{minipage}
\end{table}

\section{More Ablation Studies}
\label{sec:add_ablation}

\subsection{The Number of Canonical Joint Slots}
Table~\ref{tab:ablation_slot} shows the impact of varying the number of Canonical Joint Slots (\texttt{CJS}), with $J_c \in \{0, 5, 10, 15\}$. Introducing \texttt{CJS} consistently improves performance over the variant without canonical slots, confirming the effectiveness of our sensor-unified canonical space. The best overall performance is achieved at $J_c=5$, which improves NTU-60 \citep{NTU60}, NTU-120 \citep{NTU120}, PKU-MMD \citep{PKU}, NW-UCLA \citep{NWUCLA}, and UT-Kinect \citep{UTKinect} by 9.3, 7.8, 13.1, 6.2, and 6.0\%-point over the $J_c=0$ baseline, respectively. When the number of slots is further increased to 10 or 15, performance gradually declines on most benchmarks. We believe this is because too many canonical slots \textit{reduce the representation bottleneck effect} and \textit{introduce redundancy across slots}, making the canonical space less compact and less effective for cross-sensor semantic alignment. This suggests that a small set of canonical slots is sufficient to capture shared human motion semantics while preserving transferability across heterogeneous sensors.

\begin{table}[t]
    \scriptsize
    \centering
    \caption{Ablation study on the number of Canonical Joint Slots (\texttt{CJS}). We report linear evaluation accuracy (\%) on 25-joint and 20-joint benchmarks, together with inference GFLOPs.}
    \label{tab:ablation_slot}
    \resizebox{0.7\columnwidth}{!}{%
    \def\arraystretch{1.0}
    \setlength{\tabcolsep}{4pt}
    \begin{tabular}{ccccccc}
        \toprule
        \textbf{Slots} & \multicolumn{3}{c}{\textbf{25-Joint Datasets}} & \multicolumn{2}{c}{\textbf{20-Joint Datasets}} & \textbf{GFLOPs} \\
        \cmidrule(lr){1-1} \cmidrule(lr){2-4} \cmidrule(lr){5-6} \cmidrule(lr){7-7}
        \texttt{CJS} & NTU-60 & NTU-120 & PKU-MMD & NW-UCLA & UT-Kinect & Inference \\
        \midrule
        0 & 76.4 & 71.4 & 52.3 & 86.7 & 91.0 & 3.60 \\
        \rowcolor{yellow!15}
        5 & 85.7 & 79.2 & 65.4 & 92.9 & 97.0 & 4.30 \\
        10 & 85.5 & 79.0 & 64.5 & 92.7 & 98.0 & 5.41 \\
        15 & 85.3 & 78.9 & 63.5 & 91.8 & 97.0 & 6.55 \\
        \bottomrule
    \end{tabular}}
\end{table}

\begin{figure*}[t]
  \centering
  \includegraphics[width=0.9\textwidth]{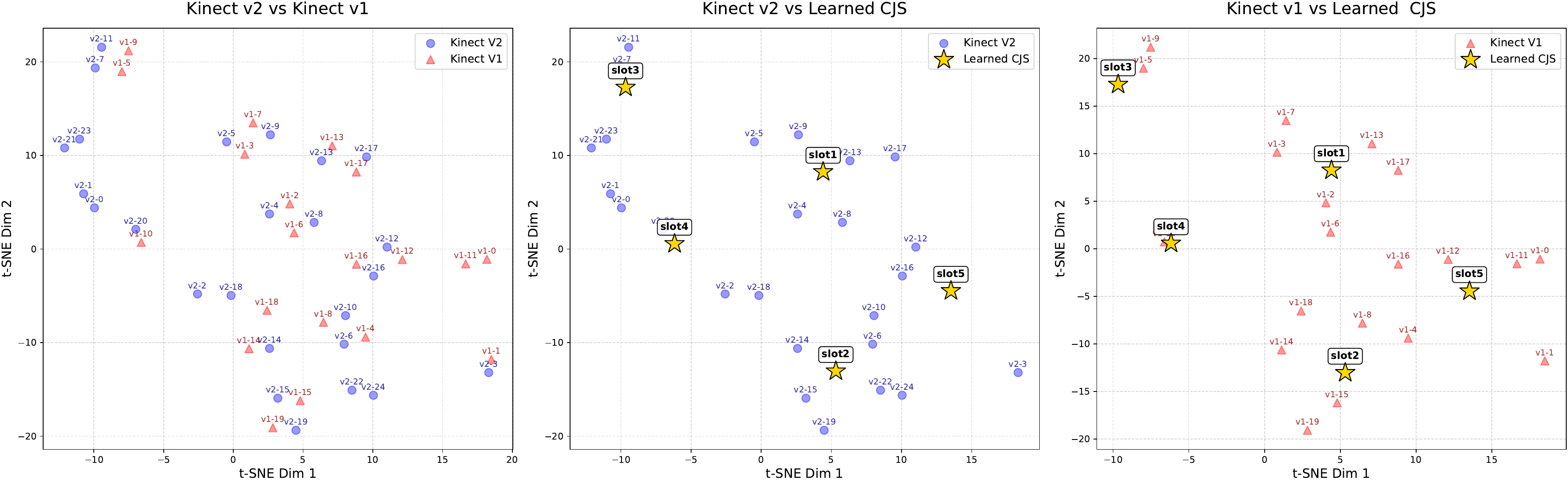}
  \caption{t-SNE visualization of Canonical Joint Slot assignments. The five slots autonomously converge toward the semantic centers of distinct anatomical clusters (torso and four limbs), consistent with the optimal $J_c=5$ in Table~\ref{tab:ablation_slot}.}
  \label{fig:cjs}
\end{figure*}

\subsection{Flexibility of the Slot Count}
\label{sec:jc_flex}
Beyond resolving joint-count discrepancy, the slot count $J_c$ endows SOfA with a tunable output resolution. By scaling $J_c$, the model generates a standardized representation tailored to specific downstream needs---ranging from compact prompts for motion diffusion, to richer structural inputs for vision--language models (VLMs), to high-level features for action recognition. As quantified in Table~\ref{tab:ablation_slot}, small slot counts already capture the shared human-motion semantics, while overly large counts dilute the bottleneck and introduce redundancy. This flexibility establishes SOfA as a versatile skeletal interface that adapts its output density to the computational and semantic requirements of diverse applications, without any architectural change.

\subsection{Interpretability of Canonical Joint Slots}
\label{sec:cjs_interp}
Using \texttt{CJS} to resolve cross-sensor topological gaps is a novel design, and we find the learned slots to be semantically meaningful rather than a generic bottleneck. Fig.~\ref{fig:cjs} visualizes a t-SNE embedding of the slot assignments: the slots autonomously converge toward the semantic centers of distinct anatomical clusters across diverse datasets. If \texttt{CJS} were merely a generic compression bottleneck, its optimal size would be arbitrary; yet Table~\ref{tab:ablation_slot} shows performance peaks precisely at five slots. This natural alignment with the five major regions of the human body (torso and four limbs) quantitatively indicates that \texttt{CJS} extracts an anatomically meaningful space rather than performing arbitrary data compression.

\begin{figure*}[t]
  \centering
  \includegraphics[width=0.7\textwidth]{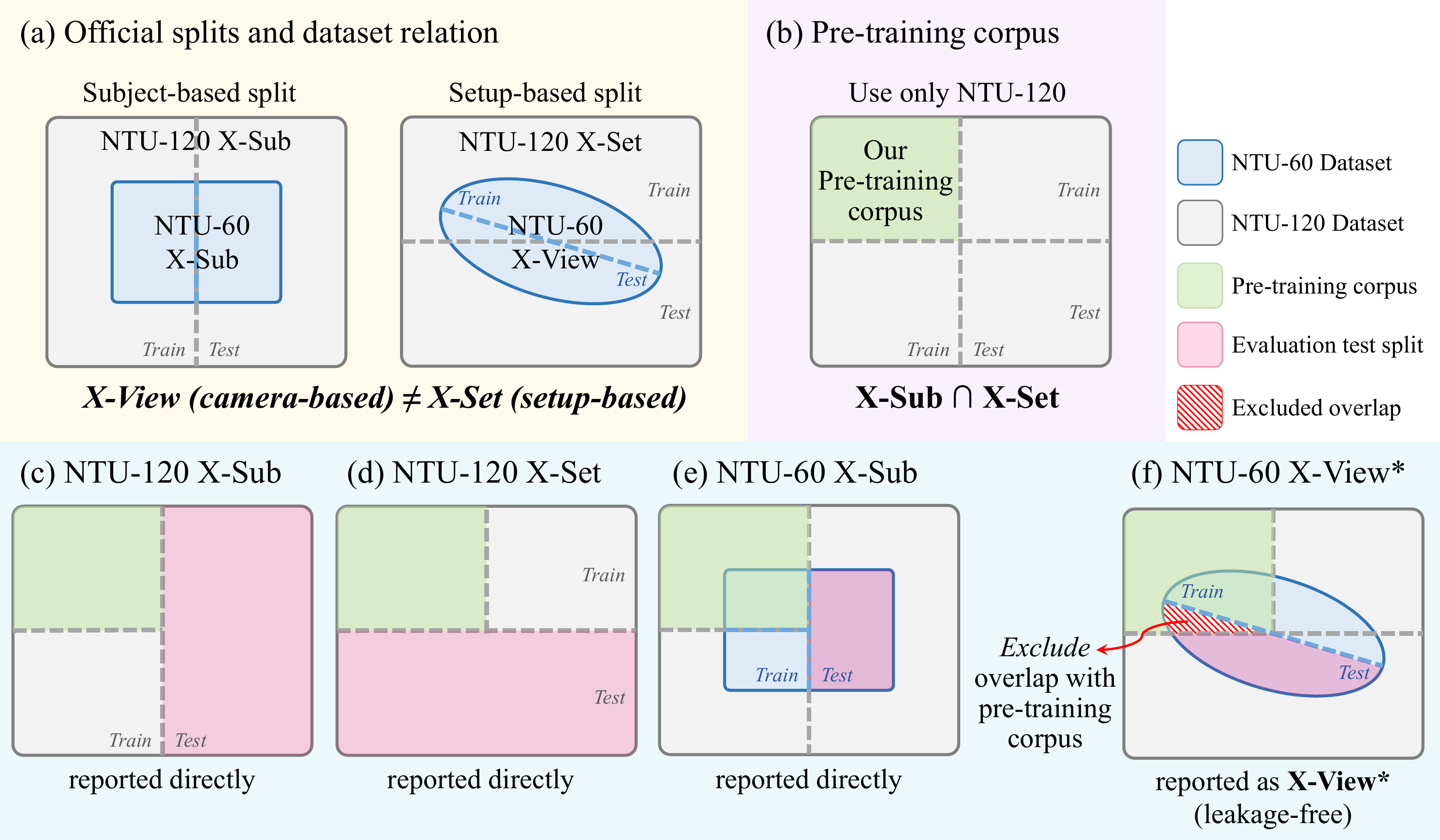}
    \caption{\textbf{Construction of the NTU pre-training corpus and evaluation sets.} NTU-120 contains all NTU-60 sequences. NTU-60 and NTU-120 use the same X-Sub assignment for all NTU-60 subjects. However, NTU-60 X-View splits samples by camera ID, while NTU-120 X-Set splits them by setup ID. (b) Pre-training uses only sequences assigned to training by both NTU-120 X-Sub and X-Set. (c--e) The official NTU-120 X-Sub, NTU-120 X-Set, and NTU-60 X-Sub test sets are used directly because they do not overlap with the pre-training corpus. (f) For NTU-60 X-View, overlapping test sequences are removed before evaluation, and the remaining leakage-free subset is denoted X-View$^{*}$.}
  \label{fig:leakage}
\end{figure*}

\begin{table*}[t]
    \scriptsize
    \setlength\tabcolsep{4pt}
    \centering
    \caption{Summary of the ten heterogeneous 3D skeleton datasets. Datasets are grouped by tracking protocol and skeletal topology; the 15-joint protocols are reserved exclusively for unseen evaluation.}
    \label{tab:dataset_summary}
    \resizebox{0.9\textwidth}{!}{
    \def\arraystretch{1.0}
    \begin{tabular}{l|c|c|c|c|l}
        \toprule
        \rowcolor{yellow!15}
        \textbf{Dataset} & \textbf{Tracker} & \textbf{Joints} & \textbf{Classes} & \textbf{Samples} & \textbf{Evaluation Setting} \\
        \midrule
        NTU-60 \citep{NTU60} & Kinect v2 & 25 & 60 & 56,880 & Pre-train \& Downstream \\
        NTU-120 \citep{NTU120} & Kinect v2 & 25 & 120 & 114,480 & Pre-train \& Downstream \\
        PKU-MMD \citep{PKU} & Kinect v2 & 25 & 51 & 7,096 & Pre-train \& Downstream \\
        ETRI-Act \citep{ETRI} & Kinect v2 & 25 & 55 & 112,620 & Pre-train \& Downstream \\
        ETRI-LivingLab \citep{ETRI} & Kinect v2 & 25 & 55 & 8,605 & Pre-train \& Downstream \\
        \midrule
        MSR-Action3D \citep{MSRAction3} & Kinect v1 & 20 & 20 & 567 & Pre-train \& Downstream \\
        NW-UCLA \citep{NWUCLA} & Kinect v1 & 20 & 10 & 1,494 & Pre-train \& Downstream \\
        UT-Kinect \citep{UTKinect} & Kinect v1 & 20 & 10 & 199 & Pre-train \& Downstream \\
        \midrule
        SBU-Inter \citep{SBU} & Custom Tracker 1 & 15 & 8 & 282 & Unseen (OOD) \\
        Florence \citep{Florence} & Custom Tracker 2 & 15 & 9 & 215 & Unseen (OOD) \\
        \bottomrule
    \end{tabular}}
\end{table*}

\section{Pre-training Corpus and Evaluation Protocol for NTU-60/120}
\label{sec:leakage}

\paragraph{Why special handling is needed.}

Previous skeleton representation methods are sensor-specific and train a separate model for each benchmark. In contrast, SOfA is the first sensor-unified approach that pre-trains a single encoder on a combined skeleton corpus. When constructing this corpus, NTU-60~\cite{NTU60} and NTU-120~\cite{NTU120} require special handling because NTU-120 is an extension of NTU-60 and contains all NTU-60 sequences.

\paragraph{Official protocols and splits.}
NTU-60 contains 56,880 raw action sequences from 60 classes and 40 subjects, whereas NTU-120 extends it to 114,480 sequences, 120 classes, and 106 subjects. Both datasets provide 3D skeleton sequences with 25 body joints captured using Kinect v2 sensors.

NTU-60 defines two official evaluation protocols: X-Sub and X-View. X-Sub divides its 40 subjects into 20 training and 20 test subjects, producing 40,320 training samples and 16,560 test samples. X-View splits samples by camera ID: cameras 2 and 3 are used for training, while camera 1 is used for testing. The resulting sets contain 37,920 training samples and 18,960 test samples. Likewise, NTU-120 defines two official evaluation protocols, X-Sub and X-Set: both datasets use X-Sub, but their second protocols differ. NTU-120 X-Sub divides the 106 subjects into 53 training and 53 test subjects, while X-Set uses the 16 even-numbered setups for training and the 16 odd-numbered setups for testing.

For the original 40 subjects, NTU-60 X-Sub and NTU-120 X-Sub use exactly the same train--test assignment. In other words, a subject used for training in NTU-60 X-Sub is also used for training in NTU-120 X-Sub, and the same holds for testing. The X-View and X-Set splits follow different rules. X-View splits sequences by camera ID, while X-Set splits them by setup ID. Therefore, the same sequence can belong to the X-Set training set but the X-View test set. Figure~\ref{fig:leakage} gives an overview of these dataset relations and official splits.

\paragraph{Pre-training corpus.} For NTU-60 and NTU-120, we use only NTU-120 during pre-training because it already contains all NTU-60 sequences. We keep only the sequences assigned to training by both NTU-120 protocols \textbf{(X-Sub $\cap$ X-Set, i.e., 25,053, only 22\% of NTU-120)}. If either X-Sub or X-Set assigns a sequence to testing, that sequence is not used for pre-training. Figure~\ref{fig:leakage}(b) illustrates this construction.

\paragraph{Evaluation on official test sets.} The official NTU-120 X-Sub and X-Set test sets can be used directly because none of their test sequences are included in the pre-training corpus. The official NTU-60 X-Sub test set can also be used directly. For the 40 subjects in NTU-60, NTU-60 and NTU-120 use the same 20 training subjects and the same 20 test subjects under X-Sub. Figure~\ref{fig:leakage}(c--e) illustrates these three cases.

\paragraph{NTU-60 X-View$^{*}$.} NTU-60 X-View needs an additional check in our sensor-unified setting. The official X-View split is valid when NTU-60 is used alone; the overlap appears only because our pre-training also uses NTU-120. X-View splits sequences by camera ID, while NTU-120 X-Set splits them by setup ID. Because these rules differ, some sequences in the official X-View test set are also included in our pre-training corpus. Before evaluation, we compare the official X-View test set with the pre-training corpus using sequence IDs and remove every overlapping sequence. We report the remaining leakage-free test set as X-View$^{*}$. Table 2 in the main paper reports accuracy on this set. This filtering is completed before evaluation and does not depend on model predictions. Figure~\ref{fig:leakage}(f) illustrates this process. We remove \textbf{5,392} sequences that also appear in the pre-training corpus, leaving \textbf{13,568} sequences in X-View$^{*}$.

\begin{figure}[t]
  \centering
  \includegraphics[width=\columnwidth]{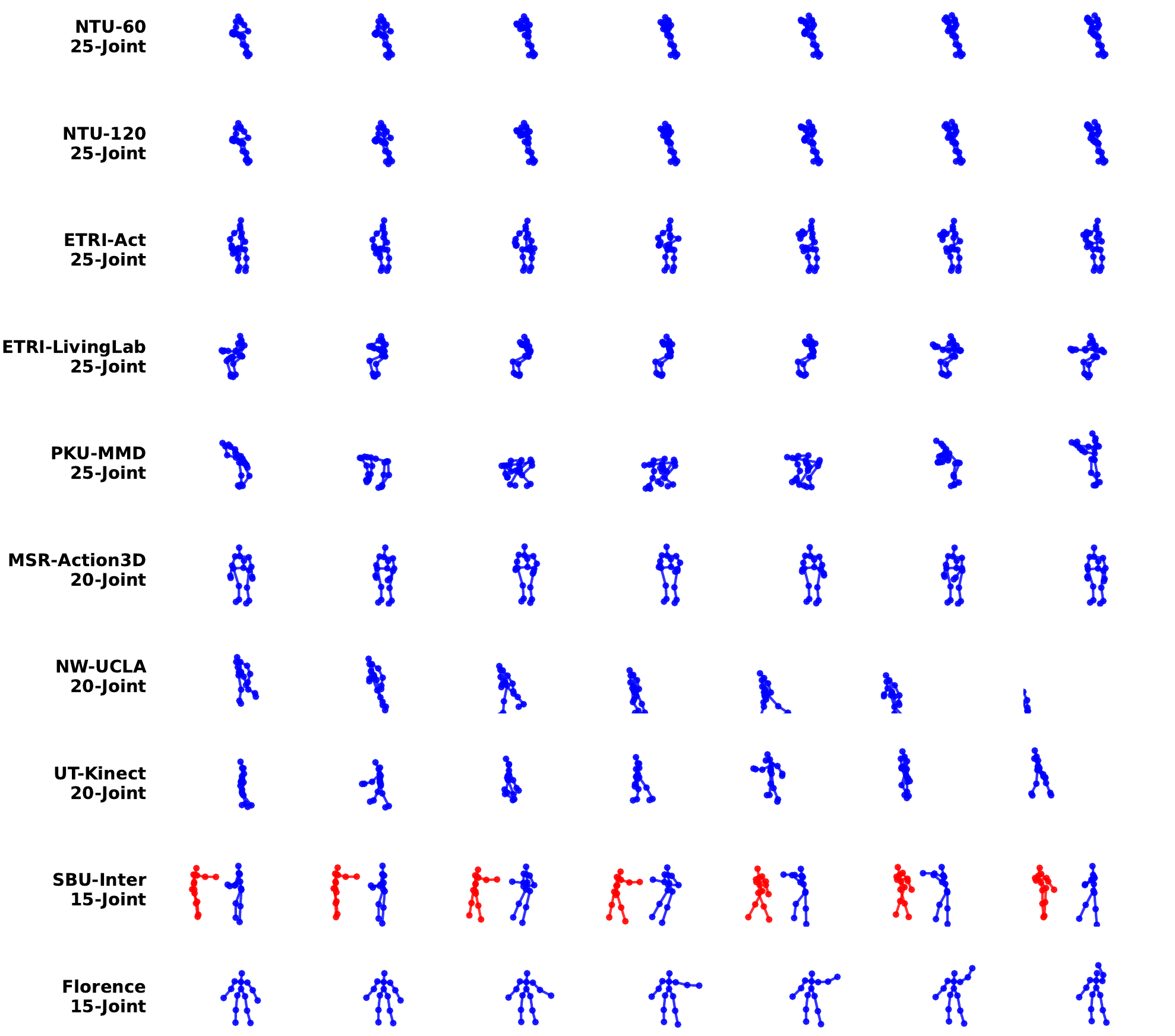}
  \caption{Representative normalized skeleton samples from the ten heterogeneous datasets. Despite variations in joint count, tracking protocol, and acquisition environment, the sequences are consistently aligned to a shared hip-centered coordinate frame.}
  \label{fig:dataset_vis1}
\end{figure}

\begin{figure}[t]
  \centering
  \includegraphics[width=\columnwidth]{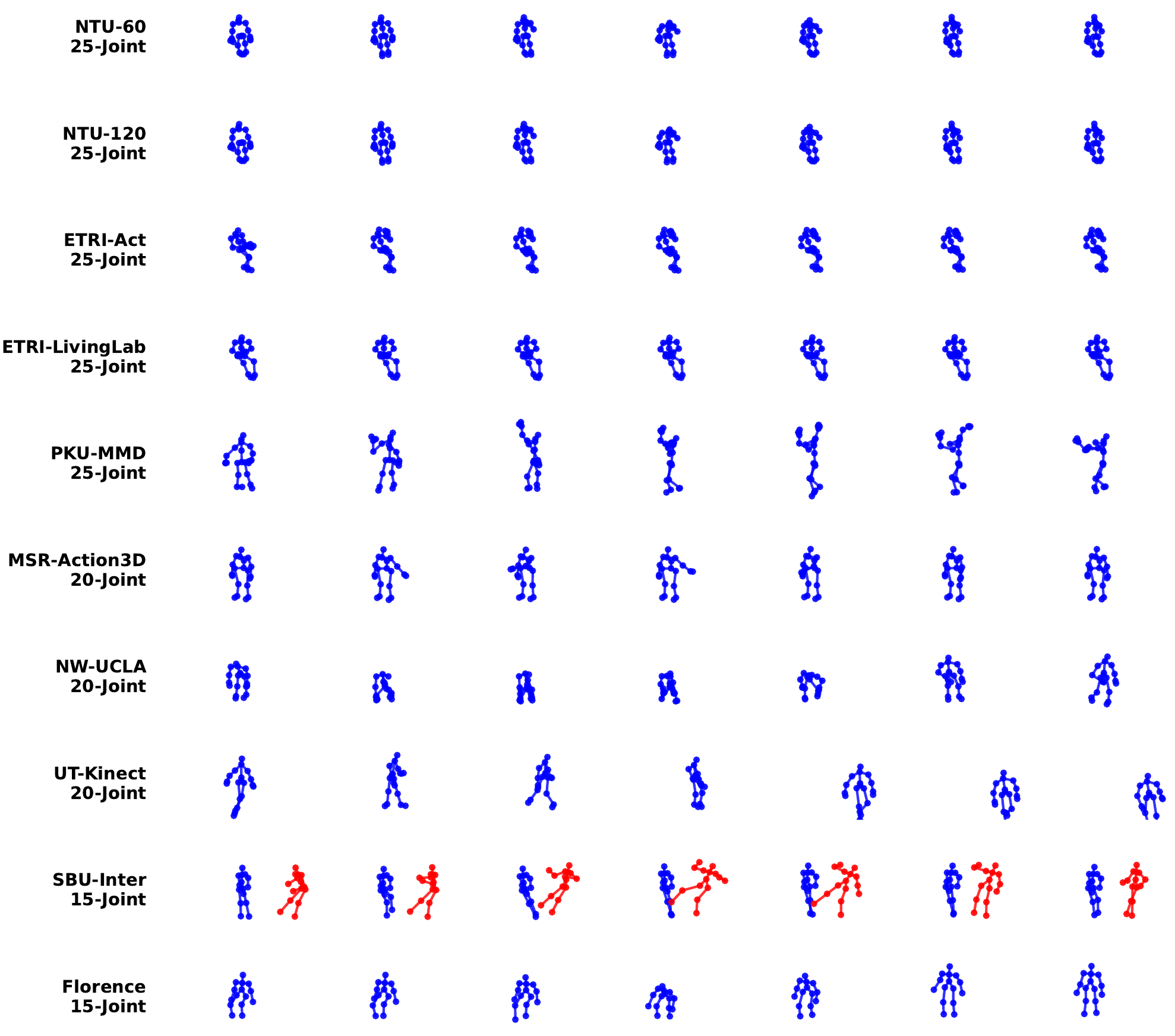}
  \caption{Additional normalized skeleton examples from the ten heterogeneous datasets, illustrating that our preprocessing preserves dataset-specific motion patterns while reducing cross-dataset geometric discrepancies.}
  \label{fig:dataset_vis2}
\end{figure}

\begin{figure}[t]
  \centering
  \includegraphics[width=\columnwidth]{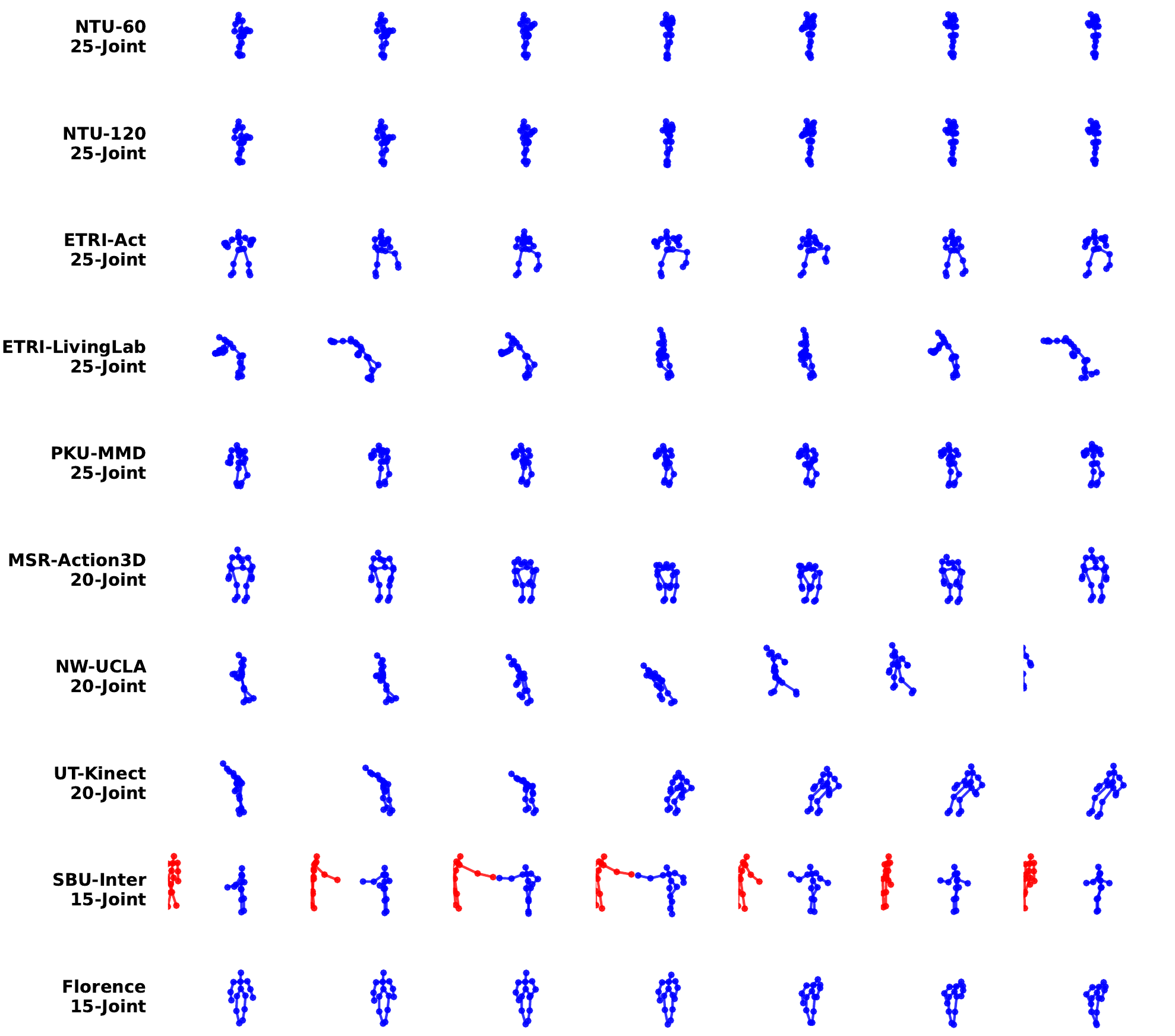}
  \caption{Additional qualitative visualization of normalized skeletons across the ten datasets, including unseen 15-joint protocols. The unified preprocessing remains effective under substantial topological variation and interaction-specific layouts.}
  \label{fig:dataset_vis3}
\end{figure}


\section{Datasets and Preprocessing}
\label{sec:dataset_detail}

\subsection{Datasets}
Table~\ref{tab:dataset_summary} summarizes the ten heterogeneous 3D skeleton datasets used in our framework. The practical heterogeneity among datasets arises from differences in tracking protocols, joint indexing conventions, anatomical coverage, and coordinate scales, leading to substantially different skeletal topologies even when some datasets originate from similar hardware families. In our benchmark design, the pre-training corpus consists of eight datasets with seen 20-joint and 25-joint protocols, while two 15-joint datasets are strictly reserved for unseen evaluation, testing whether SOfA generalizes beyond the skeletal structures observed during pre-training. The unseen 15-joint protocols are not merely lower-dimensional versions of the seen datasets but follow different joint layouts and indexing systems, making transfer non-trivial. To further illustrate these structural differences, Fig.~\ref{fig:dataset_vis1}, Fig.~\ref{fig:dataset_vis2}, and Fig.~\ref{fig:dataset_vis3} visualize representative normalized skeletons from each dataset after preprocessing.

\subsection{Preprocessing Details}
To reduce geometric inconsistencies across heterogeneous sources, we apply a unified preprocessing pipeline before pre-training and downstream evaluation. First, all coordinates are converted to meters ($m$) when necessary. Next, each sequence is translated to a shared semantic origin defined from the first frame. For datasets with an explicit ``hip'' joint, we use that joint as the origin; for 15-joint datasets such as SBU-Inter \citep{SBU} and Florence \citep{Florence}, where no explicit ``hip'' joint is available, we define a virtual hip as the midpoint between the left and right hip joints. All coordinates are then translated by the first-frame origin of the main actor, yielding a consistent hip-centered coordinate frame. We additionally align the vertical body axis to the $Y$-axis so that skeletons from different tracking protocols share a common upright orientation. Finally, we remove invalid samples, including sequences containing NaN or Inf values, all-zero skeletons, and samples with extreme coordinate outliers beyond a predefined threshold. Table~\ref{tab:coord_stats} reports per-axis coordinate statistics after preprocessing. Despite the large variation in topology and acquisition settings, the processed datasets exhibit comparable spatial ranges, showing that our standardization successfully neutralizes cross-dataset statistical discrepancies.

\begin{table}[t]
    \scriptsize
    \setlength\tabcolsep{6pt}
    \centering
    \caption{Per-axis coordinate statistics of the ten unified datasets after preprocessing (in meters). The heterogeneous datasets are consistently aligned to a shared hip-centered coordinate frame with comparable spatial ranges.}
    \label{tab:coord_stats}
    \resizebox{0.65\columnwidth}{!}{
    \def\arraystretch{1.0}
    \begin{tabular}{l|ccccc}
        \toprule
        \textbf{Dataset} & \textbf{Axis} & \textbf{Min} & \textbf{Max} & \textbf{Mean} & \textbf{Median} \\
        \midrule
        \multicolumn{6}{l}{\cellcolor{yellow!15}\textbf{25-Joint Protocols (Pre-train \& Downstream)}} \\
        \midrule
        \multirow{3}{*}{NTU-60 \citep{NTU60}}
        & X & -5.3128 & 5.2244 & -0.0060 & -0.0022 \\
        & Y & -2.7981 & 2.4020 & 0.0414 & 0.0338 \\
        & Z & -4.8746 & 3.8188 & -0.0657 & -0.0536 \\
        \midrule
        \multirow{3}{*}{NTU-120 \citep{NTU120}}
        & X & -5.3128 & 5.2244 & 0.0012 & -0.0012 \\
        & Y & -2.7981 & 2.4369 & 0.0422 & 0.0375 \\
        & Z & -4.8746 & 4.5483 & -0.0703 & -0.0515 \\
        \midrule
        \multirow{3}{*}{PKU-MMD \citep{PKU}}
        & X & -2.9480 & 2.2948 & 0.0062 & 0.0021 \\
        & Y & -1.7711 & 2.5059 & 0.1165 & 0.1512 \\
        & Z & -4.2423 & 4.6754 & -0.0492 & -0.0420 \\
        \midrule
        \multirow{3}{*}{ETRI-Act \citep{ETRI}}
        & X & -3.0403 & 3.3122 & 0.0158 & 0.0079 \\
        & Y & -2.9967 & 2.4528 & 0.1052 & 0.1049 \\
        & Z & -3.7749 & 2.9988 & -0.0910 & -0.0635 \\
        \midrule
        \multirow{3}{*}{ETRI-LivingLab \citep{ETRI}}
        & X & -3.1178 & 2.2961 & 0.0071 & 0.0030 \\
        & Y & -1.8460 & 2.5984 & 0.1005 & 0.1201 \\
        & Z & -3.0634 & 2.5608 & -0.0704 & -0.0564 \\
        \midrule
        \multicolumn{6}{l}{\cellcolor{yellow!15}\textbf{20-Joint Protocols (Pre-train \& Downstream)}} \\
        \midrule
        \multirow{3}{*}{MSR-Action3D \citep{MSRAction3}}
        & X & -0.8821 & 1.0665 & -0.0012 & -0.0031 \\
        & Y & -1.8576 & 1.1964 & -0.1543 & -0.0096 \\
        & Z & -1.1771 & 2.7745 & -0.0059 & 0.0132 \\
        \midrule
        \multirow{3}{*}{NW-UCLA \citep{NWUCLA}}
        & X & -3.1459 & 3.1226 & -0.0090 & -0.0079 \\
        & Y & -1.7658 & 1.3540 & -0.1120 & -0.0121 \\
        & Z & -2.6875 & 2.3884 & -0.0016 & 0.0026 \\
        \midrule
        \multirow{3}{*}{UT-Kinect \citep{UTKinect}}
        & X & -1.8810 & 2.1565 & -0.0431 & -0.0084 \\
        & Y & -1.3734 & 1.2126 & -0.1029 & -0.0028 \\
        & Z & -2.2119 & 2.2332 & -0.0390 & -0.0062 \\
        \midrule
        \multicolumn{6}{l}{\cellcolor{yellow!15}\textbf{15-Joint Protocols (Unseen)}} \\
        \midrule
        \multirow{3}{*}{SBU-Inter \citep{SBU}}
        & X & -2.3407 & 2.1743 & -0.0727 & -0.0425 \\
        & Y & -1.2912 & 1.0371 & 0.0211 & 0.0989 \\
        & Z & -1.0182 & 0.9315 & 0.0163 & 0.0242 \\
        \midrule
        \multirow{3}{*}{Florence \citep{Florence}}
        & X & -0.7025 & 0.7916 & 0.0071 & 0.0116 \\
        & Y & -1.6579 & 1.1580 & -0.0384 & 0.0046 \\
        & Z & -0.8494 & 0.7731 & -0.0344 & -0.0145 \\
        \bottomrule
    \end{tabular}}
\end{table}

\section{Implementation Details}
\label{sec:implement}

\noindent\textbf{Encoder architecture.}
The backbone encoder $g$ is a customized Vision Transformer (ViT) \citep{ViT} tailored for skeleton representation learning. It consists of 8 Transformer layers with a hidden dimension of $D=256$ and 8 attention heads. To alleviate attention artifacts and reduce the risk of feature collapse, we incorporate 4 register tokens ($N_{\text{reg}}=4$) \citep{Register}, which absorb outlier features during self-attention. Both the student network $f_{\bm\theta}$ and the teacher network $f_{\bm\phi}$ employ identical 3-layer MLP projection heads, $h_{\texttt{Cano}}$ and $h_{\texttt{DINO}}$, with a hidden dimension of 2,048 and a bottleneck dimension of 256. These heads project the encoded features into a high-dimensional prototype space with $K=65{,}536$ prototypes, enabling the model to capture diverse motion patterns. Accordingly, the probability distributions $\mathbf{p}$ in the two distillation losses are $K$-dimensional.

\noindent\textbf{Training objectives.}
The framework is optimized with a dual-objective loss combining the Canonical Slot Reconstruction loss $\mathcal{L}_{\text{Cano}}$ and the DINO-based distillation loss $\mathcal{L}_{\text{DINO}}$ \citep{DINO} with an equal weighting factor ($\lambda=1$). To further regularize the embedding space and encourage features to be uniformly distributed on the unit hypersphere, we additionally employ the KoLeo regularization loss \citep{DINOv2} with weight 0.1, and apply the Sinkhorn--Knopp algorithm \citep{DINOv2} for prototype-assignment regularization, which prevents mode collapse and promotes balanced usage of the prototype space.

\noindent\textbf{Temporal sampling.}
To enable batch-wise training across datasets with different sequence lengths and frame rates, we standardize the temporal resolution of all inputs. For each raw sequence, we randomly crop a temporal window spanning 50\% to 100\% of the original duration and uniformly resample it to a fixed length of $T=64$ frames, yielding a unified input tensor $\mathbf{X} \in \mathbb{R}^{T \times J_{\text{max}} \times 3}$.

\noindent\textbf{Augmentation and masking.}
To improve robustness and generalization, we apply rotation, scaling, spatial flipping, and axis dropout, each with probability $p=0.5$. Our masking serves two purposes. First, for general feature robustness, we apply joint-wise masking as a data augmentation with probability $p=0.5$. Second, for the Canonical Slot Reconstruction objective ($\mathcal{L}_{\text{Cano}}$), we always apply masking ($p=1.0$) with a high masking ratio of 90\%. By exposing the student network to only a severely corrupted subset of the input, the model is encouraged to internalize human kinematics and infer the missing information required to reconstruct the Canonical Joint Slots from sparse visible cues.

%% file: iclr2027_conference.bib
@String(CVPR  = {IEEE Conf. Comput. Vis. Pattern Recog.})

@String(ICCV  = {Int. Conf. Comput. Vis.})

@String(ECCV  = {Eur. Conf. Comput. Vis.})

@String(AAAI  = {AAAI})

@String(CVPR  = {CVPR})

@String(ICCV  = {ICCV})

@String(ECCV  = {ECCV})

@article{Review1,
  title={Human action recognition from various data modalities: A review},
  author={Sun, Zehua and Ke, Qiuhong and Rahmani, Hossein and Bennamoun, Mohammed and Wang, Gang and Liu, Jun},
  journal={IEEE transactions on pattern analysis and machine intelligence},
  volume={45},
  number={3},
  pages={3200--3225},
  year={2022},
  publisher={IEEE}
}

@article{Review2,
  title={Self-Supervised Skeleton-Based Action Representation Learning: A Benchmark and Beyond: J. Zhang et al.},
  author={Zhang, Jiahang and Lin, Lilang and Yang, Shuai and Liu, Jiaying},
  journal={International Journal of Computer Vision},
  volume={134},
  number={1},
  pages={38},
  year={2026},
  publisher={Springer}
}

@article{Review3,
  title={Human action recognition and prediction: A survey},
  author={Kong, Yu and Fu, Yun},
  journal={International Journal of Computer Vision},
  volume={130},
  number={5},
  pages={1366--1401},
  year={2022},
  publisher={Springer}
}

@inproceedings{AMR,
  title={Exploring Adaptive Masked Reconstruction for Self-Supervised Skeleton-Based Action Recognition},
  author={Sun, Shengkai and Cheng, Zhiyong and Zhang, Zefan and Dong, Jianfeng and Li, Zhihui and Wang, Meng},
  booktitle={Proceedings of the IEEE/CVF Conference on Computer Vision and Pattern Recognition},
  pages={13974--13983},
  year={2026}
}

@article{CrosSCLR,
  title={3d human action representation learning via cross-view consistency pursuit},
  author={Li, Linguo and Wang, Minsi and Ni, Bingbing and Wang, Hang and Yang, Jiancheng and Zhang, Wenjun},
  journal={arXiv preprint arXiv:2104.14466},
  year={2021}
}

@inproceedings{GL-Transformer,
  title={Global-local motion transformer for unsupervised skeleton-based action learning},
  author={Kim, Boeun and Chang, Hyung Jin and Kim, Jungho and Choi, Jin Young},
  booktitle={European conference on computer vision},
  pages={209--225},
  year={2022},
  organization={Springer}
}

@inproceedings{MacDiff,
  title={Macdiff: Unified skeleton modeling with masked conditional diffusion},
  author={Wu, Lehong and Lin, Lilang and Zhang, Jiahang and Ma, Yiyang and Liu, Jiaying},
  booktitle={European Conference on Computer Vision},
  pages={110--128},
  year={2024},
  organization={Springer}
}

@inproceedings{IGM,
  title={Idempotent unsupervised representation learning for skeleton-based action recognition},
  author={Lin, Lilang and Wu, Lehong and Zhang, Jiahang and Liu, Jiaying},
  booktitle={European Conference on Computer Vision},
  pages={75--92},
  year={2024},
  organization={Springer}
}

@inproceedings{USDRL,
  title={Usdrl: Unified skeleton-based dense representation learning with multi-grained feature decorrelation},
  author={Weng, Wanjiang and Wang, Hongsong and Wang, Junbo and He, Lei and Xie, Guo-Sen},
  booktitle={Proceedings of the AAAI Conference on Artificial Intelligence},
  volume={39},
  pages={8332--8340},
  year={2025}
}

@inproceedings{HSP,
  title={Heterogeneous skeleton-based action representation learning},
  author={Wang, Hongsong and Ma, Xiaoyan and Kuang, Jidong and Gui, Jie},
  booktitle={Proceedings of the Computer Vision and Pattern Recognition Conference},
  pages={19154--19164},
  year={2025}
}

@inproceedings{CPM,
  title={Contrastive positive mining for unsupervised 3d action representation learning},
  author={Zhang, Haoyuan and Hou, Yonghong and Zhang, Wenjing and Li, Wanqing},
  booktitle={European Conference on Computer Vision},
  pages={36--51},
  year={2022},
  organization={Springer}
}

@inproceedings{CMD,
  title={Cmd: Self-supervised 3d action representation learning with cross-modal mutual distillation},
  author={Mao, Yunyao and Zhou, Wengang and Lu, Zhenbo and Deng, Jiajun and Li, Houqiang},
  booktitle={European Conference on Computer Vision},
  pages={734--752},
  year={2022},
  organization={Springer}
}

@inproceedings{AimCLR,
  title={Contrastive learning from extremely augmented skeleton sequences for self-supervised action recognition},
  author={Guo, Tianyu and Liu, Hong and Chen, Zhan and Liu, Mengyuan and Wang, Tao and Ding, Runwei},
  booktitle={Proceedings of the AAAI Conference on Artificial Intelligence},
  volume={36},
  pages={762--770},
  year={2022}
}

@inproceedings{RVTCLR,
  title={Modeling the relative visual tempo for self-supervised skeleton-based action recognition},
  author={Zhu, Yisheng and Han, Hu and Yu, Zhengtao and Liu, Guangcan},
  booktitle={Proceedings of the IEEE/CVF International Conference on Computer Vision},
  pages={13913--13922},
  year={2023}
}

@inproceedings{PTSL,
  title={Self-supervised action representation learning from partial spatio-temporal skeleton sequences},
  author={Zhou, Yujie and Duan, Haodong and Rao, Anyi and Su, Bing and Wang, Jiaqi},
  booktitle={Proceedings of the AAAI conference on artificial intelligence},
  volume={37},
  pages={3825--3833},
  year={2023}
}

@article{HYSP,
  title={Hyperbolic self-paced learning for self-supervised skeleton-based action representations},
  author={Franco, Luca and Mandica, Paolo and Munjal, Bharti and Galasso, Fabio},
  journal={arXiv preprint arXiv:2303.06242},
  year={2023}
}

@inproceedings{HaLP,
  title={Halp: Hallucinating latent positives for skeleton-based self-supervised learning of actions},
  author={Shah, Anshul and Roy, Aniket and Shah, Ketul and Mishra, Shlok and Jacobs, David and Cherian, Anoop and Chellappa, Rama},
  booktitle={Proceedings of the IEEE/CVF Conference on Computer Vision and Pattern Recognition},
  pages={18846--18856},
  year={2023}
}

@inproceedings{ActCLR,
  title={Actionlet-Dependent Contrastive Learning for Unsupervised Skeleton-Based Action Recognition},
  author={Lin, Lilang and Zhang, Jiahang and Liu, Jiaying},
  booktitle={Proceedings of the IEEE/CVF Conference on Computer Vision and Pattern Recognition},
  pages={2363--2372},
  year={2023}
}

@inproceedings{SkeletonMAE,
  title={Skeletonmae: Spatial-temporal masked autoencoders for self-supervised skeleton action recognition},
  author={Wu, Wenhan and Hua, Yilei and Zheng, Ce and Wu, Shiqian and Chen, Chen and Lu, Aidong},
  booktitle={2023 IEEE international conference on multimedia and expo workshops (ICMEW)},
  pages={224--229},
  year={2023},
  organization={IEEE}
}

@inproceedings{MAMP,
  title={Masked motion predictors are strong 3d action representation learners},
  author={Mao, Yunyao and Deng, Jiajun and Zhou, Wengang and Fang, Yao and Ouyang, Wanli and Li, Houqiang},
  booktitle={Proceedings of the IEEE/CVF International Conference on Computer Vision},
  pages={10181--10191},
  year={2023}
}

@inproceedings{SJEPA,
  title={S-jepa: A joint embedding predictive architecture for skeletal action recognition},
  author={Abdelfattah, Mohamed and Alahi, Alexandre},
  booktitle={European Conference on Computer Vision},
  pages={367--384},
  year={2024},
  organization={Springer}
}

@inproceedings{GFP,
  title={Towards Efficient General Feature Prediction in Masked Skeleton Modeling},
  author={Sun, Shengkai and Zhang, Zefan and Dong, Jianfeng and Cheng, Zhiyong and Chang, Xiaojun and Wang, Meng},
  booktitle={Proceedings of the IEEE/CVF International Conference on Computer Vision},
  pages={12212--12221},
  year={2025}
}

@inproceedings{LongTGAN,
  title={Unsupervised representation learning with long-term dynamics for skeleton based action recognition},
  author={Zheng, Nenggan and Wen, Jun and Liu, Risheng and Long, Liangqu and Dai, Jianhua and Gong, Zhefeng},
  booktitle={Proceedings of the AAAI conference on artificial intelligence},
  volume={32},
  year={2018}
}

@inproceedings{PandC,
  title={Predict \& cluster: Unsupervised skeleton based action recognition},
  author={Su, Kun and Liu, Xiulong and Shlizerman, Eli},
  booktitle={Proceedings of the IEEE/CVF conference on computer vision and pattern recognition},
  pages={9631--9640},
  year={2020}
}

@inproceedings{ISC,
  title={Skeleton-contrastive 3D action representation learning},
  author={Thoker, Fida Mohammad and Doughty, Hazel and Snoek, Cees GM},
  booktitle={Proceedings of the 29th ACM international conference on multimedia},
  pages={1655--1663},
  year={2021}
}

@inproceedings{UmURL,
  title={Unified multi-modal unsupervised representation learning for skeleton-based action understanding},
  author={Sun, Shengkai and Liu, Daizong and Dong, Jianfeng and Qu, Xiaoye and Gao, Junyu and Yang, Xun and Wang, Xun and Wang, Meng},
  booktitle={Proceedings of the 31st ACM International Conference on Multimedia},
  pages={2973--2984},
  year={2023}
}

@inproceedings{HiCo,
  title={Hierarchical contrast for unsupervised skeleton-based action representation learning},
  author={Dong, Jianfeng and Sun, Shengkai and Liu, Zhonglin and Chen, Shujie and Liu, Baolong and Wang, Xun},
  booktitle={Proceedings of the AAAI Conference on Artificial Intelligence},
  volume={37},
  pages={525--533},
  year={2023}
}

@inproceedings{PoseC3D,
  title={Revisiting skeleton-based action recognition},
  author={Duan, Haodong and Zhao, Yue and Chen, Kai and Lin, Dahua and Dai, Bo},
  booktitle={Proceedings of the IEEE/CVF conference on computer vision and pattern recognition},
  pages={2969--2978},
  year={2022}
}

@inproceedings{CTRGCN,
  title={Channel-wise topology refinement graph convolution for skeleton-based action recognition},
  author={Chen, Yuxin and Zhang, Ziqi and Yuan, Chunfeng and Li, Bing and Deng, Ying and Hu, Weiming},
  booktitle={Proceedings of the IEEE/CVF international conference on computer vision},
  pages={13359--13368},
  year={2021}
}

@inproceedings{Skateformer,
  title={Skateformer: skeletal-temporal transformer for human action recognition},
  author={Do, Jeonghyeok and Kim, Munchurl},
  booktitle={European Conference on Computer Vision},
  pages={401--420},
  year={2024},
  organization={Springer}
}

@article{ViT,
  title={An image is worth 16x16 words: Transformers for image recognition at scale},
  author={Dosovitskiy, Alexey and Beyer, Lucas and Kolesnikov, Alexander and Weissenborn, Dirk and Zhai, Xiaohua and Unterthiner, Thomas and Dehghani, Mostafa and Minderer, Matthias and Heigold, Georg and Gelly, Sylvain and others},
  journal={arXiv preprint arXiv:2010.11929},
  year={2020}
}

@inproceedings{NTU60,
  title={Ntu rgb+ d: A large scale dataset for 3d human activity analysis},
  author={Shahroudy, Amir and Liu, Jun and Ng, Tian-Tsong and Wang, Gang},
  booktitle={Proceedings of the IEEE conference on computer vision and pattern recognition},
  pages={1010--1019},
  year={2016}
}

@article{NTU120,
  title={Ntu rgb+ d 120: A large-scale benchmark for 3d human activity understanding},
  author={Liu, Jun and Shahroudy, Amir and Perez, Mauricio and Wang, Gang and Duan, Ling-Yu and Kot, Alex C},
  journal={IEEE transactions on pattern analysis and machine intelligence},
  volume={42},
  number={10},
  pages={2684--2701},
  year={2019},
  publisher={IEEE}
}

@article{PKU,
  title={Pku-mmd: A large scale benchmark for continuous multi-modal human action understanding},
  author={Liu, Chunhui and Hu, Yueyu and Li, Yanghao and Song, Sijie and Liu, Jiaying},
  journal={arXiv preprint arXiv:1703.07475},
  year={2017}
}

@article{AdamW,
  title={Decoupled weight decay regularization},
  author={Loshchilov, Ilya and Hutter, Frank},
  journal={arXiv preprint arXiv:1711.05101},
  year={2017}
}

@article{CosineAnneal,
  title={Sgdr: Stochastic gradient descent with warm restarts},
  author={Loshchilov, Ilya and Hutter, Frank},
  journal={arXiv preprint arXiv:1608.03983},
  year={2016}
}

@inproceedings{DINO,
  title={Emerging properties in self-supervised vision transformers},
  author={Caron, Mathilde and Touvron, Hugo and Misra, Ishan and J{\'e}gou, Herv{\'e} and Mairal, Julien and Bojanowski, Piotr and Joulin, Armand},
  booktitle={Proceedings of the IEEE/CVF international conference on computer vision},
  pages={9650--9660},
  year={2021}
}

@inproceedings{CL,
  title={A simple framework for contrastive learning of visual representations},
  author={Chen, Ting and Kornblith, Simon and Norouzi, Mohammad and Hinton, Geoffrey},
  booktitle={International conference on machine learning},
  pages={1597--1607},
  year={2020},
  organization={PmLR}
}

@article{DINOv2,
  title={Dinov2: Learning robust visual features without supervision},
  author={Oquab, Maxime and Darcet, Timoth{\'e}e and Moutakanni, Th{\'e}o and Vo, Huy and Szafraniec, Marc and Khalidov, Vasil and Fernandez, Pierre and Haziza, Daniel and Massa, Francisco and El-Nouby, Alaaeldin and others},
  journal={arXiv preprint arXiv:2304.07193},
  year={2023}
}

@article{Register,
  title={Vision transformers need registers},
  author={Darcet, Timoth{\'e}e and Oquab, Maxime and Mairal, Julien and Bojanowski, Piotr},
  journal={arXiv preprint arXiv:2309.16588},
  year={2023}
}

@article{RoPE,
  title={Roformer: Enhanced transformer with rotary position embedding},
  author={Su, Jianlin and Ahmed, Murtadha and Lu, Yu and Pan, Shengfeng and Bo, Wen and Liu, Yunfeng},
  journal={Neurocomputing},
  volume={568},
  pages={127063},
  year={2024},
  publisher={Elsevier}
}

@inproceedings{MAE,
  title={Masked autoencoders are scalable vision learners},
  author={He, Kaiming and Chen, Xinlei and Xie, Saining and Li, Yanghao and Doll{\'a}r, Piotr and Girshick, Ross},
  booktitle={Proceedings of the IEEE/CVF conference on computer vision and pattern recognition},
  pages={16000--16009},
  year={2022}
}

@article{Teacher,
  title={Mean teachers are better role models: Weight-averaged consistency targets improve semi-supervised deep learning results},
  author={Tarvainen, Antti and Valpola, Harri},
  journal={Advances in neural information processing systems},
  volume={30},
  year={2017}
}

@article{Sensor,
  title={Microsoft kinect sensor and its effect},
  author={Zhang, Zhengyou},
  journal={IEEE multimedia},
  volume={19},
  number={2},
  pages={4--10},
  year={2012},
  publisher={IEEE}
}

@inproceedings{SkySense,
  title={Skysense: A multi-modal remote sensing foundation model towards universal interpretation for earth observation imagery},
  author={Guo, Xin and Lao, Jiangwei and Dang, Bo and Zhang, Yingying and Yu, Lei and Ru, Lixiang and Zhong, Liheng and Huang, Ziyuan and Wu, Kang and Hu, Dingxiang and others},
  booktitle={Proceedings of the IEEE/CVF Conference on Computer Vision and Pattern Recognition},
  pages={27672--27683},
  year={2024}
}

@inproceedings{Copernicus,
  title={Towards a unified copernicus foundation model for earth vision},
  author={Wang, Yi and Xiong, Zhitong and Liu, Chenying and Stewart, Adam J and Dujardin, Thomas and Bountos, Nikolaos Ioannis and Zavras, Angelos and Gerken, Franziska and Papoutsis, Ioannis and Leal-Taix{\'e}, Laura and others},
  booktitle={Proceedings of the IEEE/CVF International Conference on Computer Vision},
  pages={9888--9899},
  year={2025}
}

@inproceedings{SAM,
  title={Segment anything},
  author={Kirillov, Alexander and Mintun, Eric and Ravi, Nikhila and Mao, Hanzi and Rolland, Chloe and Gustafson, Laura and Xiao, Tete and Whitehead, Spencer and Berg, Alexander C and Lo, Wan-Yen and others},
  booktitle={Proceedings of the IEEE/CVF international conference on computer vision},
  pages={4015--4026},
  year={2023}
}

@article{LLM,
  title={A survey of large language models},
  author={Zhao, Wayne Xin and Zhou, Kun and Li, Junyi and Tang, Tianyi and Wang, Xiaolei and Hou, Yupeng and Min, Yingqian and Zhang, Beichen and Zhang, Junjie and Dong, Zican and others},
  journal={arXiv preprint arXiv:2303.18223},
  volume={1},
  number={2},
  pages={1--124},
  year={2023}
}

@inproceedings{PointBERT,
  title={Point-BERT: Pre-Training 3D Point Cloud Transformers with Masked Point Modeling},
  author={Yu, Xumin and Tang, Lulu and Rao, Yongming and Huang, Tiejun and Zhou, Jie and Lu, Jiwen},
  booktitle={Proceedings of the IEEE/CVF Conference on Computer Vision and Pattern Recognition (CVPR)},
  pages={19313--19322},
  year={2022}
}

@inproceedings{PointMAE,
  title={Masked Autoencoders for Point Cloud Self-supervised Learning},
  author={Pang, Yatian and Wang, Wenxiao and Tay, Francis E.H. and Liu, Wei and Tian, Yonghong and Yuan, Li},
  booktitle={European Conference on Computer Vision (ECCV)},
  year={2022}
}

@article{MaskedColor,
  title={Self-supervised 3D action representation learning with skeleton cloud colorization},
  author={Yang, Siyuan and Liu, Jun and Lu, Shijian and Hwa, Er Meng and Hu, Yongjian and Kot, Alex C},
  journal={IEEE Transactions on Pattern Analysis and Machine Intelligence},
  volume={46},
  number={1},
  pages={509--524},
  year={2023},
  publisher={IEEE}
}

@article{MCAE,
  title={Unsupervised motion representation learning with capsule autoencoders},
  author={Xu, Ziwei and Shen, Xudong and Wong, Yongkang and Kankanhalli, Mohan S},
  journal={Advances in Neural Information Processing Systems},
  volume={34},
  pages={3205--3217},
  year={2021}
}

@inproceedings{SeBiReNet,
  title={Unsupervised 3d human pose representation with viewpoint and pose disentanglement},
  author={Nie, Qiang and Liu, Ziwei and Liu, Yunhui},
  booktitle={European conference on computer vision},
  pages={102--118},
  year={2020},
  organization={Springer}
}

@inproceedings{Colorization,
  title={Skeleton cloud colorization for unsupervised 3d action representation learning},
  author={Yang, Siyuan and Liu, Jun and Lu, Shijian and Er, Meng Hwa and Kot, Alex C},
  booktitle={Proceedings of the IEEE/CVF International Conference on Computer Vision},
  pages={13423--13433},
  year={2021}
}

@inproceedings{ETRI,
  title={ETRI-activity3D: A large-scale RGB-D dataset for robots to recognize daily activities of the elderly},
  author={Jang, Jinhyeok and Kim, Dohyung and Park, Cheonshu and Jang, Minsu and Lee, Jaeyeon and Kim, Jaehong},
  booktitle={2020 IEEE/RSJ International Conference on Intelligent Robots and Systems (IROS)},
  pages={10990--10997},
  year={2020},
  organization={IEEE}
}

@inproceedings{ISTA,
  title={Interactive spatiotemporal token attention network for skeleton-based general interactive action recognition},
  author={Wen, Yuhang and Tang, Zixuan and Pang, Yunsheng and Ding, Beichen and Liu, Mengyuan},
  booktitle={2023 IEEE/RSJ International Conference on Intelligent Robots and Systems (IROS)},
  pages={7886--7892},
  year={2023},
  organization={IEEE}
}

@inproceedings{CoLSTM,
  title={Co-occurrence feature learning for skeleton based action recognition using regularized deep LSTM networks},
  author={Zhu, Wentao and Lan, Cuiling and Xing, Junliang and Zeng, Wenjun and Li, Yanghao and Shen, Li and Xie, Xiaohui},
  booktitle={Proceedings of the AAAI conference on artificial intelligence},
  volume={30},
  year={2016}
}

@inproceedings{STLSTM,
  title={Spatio-temporal lstm with trust gates for 3d human action recognition},
  author={Liu, Jun and Shahroudy, Amir and Xu, Dong and Wang, Gang},
  booktitle={European conference on computer vision},
  pages={816--833},
  year={2016},
  organization={Springer}
}

@inproceedings{VALSTM,
  title={View adaptive recurrent neural networks for high performance human action recognition from skeleton data},
  author={Zhang, Pengfei and Lan, Cuiling and Xing, Junliang and Zeng, Wenjun and Xue, Jianru and Zheng, Nanning},
  booktitle={Proceedings of the IEEE international conference on computer vision},
  pages={2117--2126},
  year={2017}
}

@article{GCA,
  title={Skeleton-based human action recognition with global context-aware attention LSTM networks},
  author={Liu, Jun and Wang, Gang and Duan, Ling-Yu and Abdiyeva, Kamila and Kot, Alex C},
  journal={IEEE Transactions on Image Processing},
  volume={27},
  number={4},
  pages={1586--1599},
  year={2017},
  publisher={IEEE}
}

@article{LSTMIRN,
  title={Interaction relational network for mutual action recognition},
  author={Perez, Mauricio and Liu, Jun and Kot, Alex C},
  journal={IEEE Transactions on Multimedia},
  volume={24},
  pages={366--376},
  year={2021},
  publisher={IEEE}
}

@inproceedings{IGFormer,
  title={Igformer: Interaction graph transformer for skeleton-based human interaction recognition},
  author={Pang, Yunsheng and Ke, Qiuhong and Rahmani, Hossein and Bailey, James and Liu, Jun},
  booktitle={European Conference on Computer Vision},
  pages={605--622},
  year={2022},
  organization={Springer}
}

@inproceedings{DiT,
  title={Scalable Diffusion Models with Transformers},
  author={Peebles, William and Xie, Saining},
  booktitle={Proceedings of the IEEE/CVF International Conference on Computer Vision (ICCV)},
  pages={4195--4205},
  year={2023}
}

@article{MMDiT,
  title={Scaling Rectified Flow Transformers for High-Resolution Image Synthesis},
  author={Esser, Patrick and Kulal, Sumith and Blattmann, Andreas and Entezari, Rahim and M{\"u}ller, Jonas and Saini, Harry and Levi, Yam and Lorenz, Dominik and Sauer, Axel and Boesel, Frederic and others},
  journal={arXiv preprint arXiv:2403.03206},
  year={2024}
}

@inproceedings{MSRAction3,
  title={Action recognition based on a bag of 3d points},
  author={Li, Wanqing and Zhang, Zhengyou and Liu, Zicheng},
  booktitle={2010 IEEE computer society conference on computer vision and pattern recognition-workshops},
  pages={9--14},
  year={2010},
  organization={IEEE}
}

@inproceedings{NWUCLA,
  title={Cross-view action modeling, learning and recognition},
  author={Wang, Jiang and Nie, Xiaohan and Xia, Yin and Wu, Ying and Zhu, Song-Chun},
  booktitle={Proceedings of the IEEE conference on computer vision and pattern recognition},
  pages={2649--2656},
  year={2014}
}

@inproceedings{UTKinect,
  title={View invariant human action recognition using histograms of 3d joints},
  author={Xia, Lu and Chen, Chia-Chih and Aggarwal, Jake K},
  booktitle={2012 IEEE computer society conference on computer vision and pattern recognition workshops},
  pages={20--27},
  year={2012},
  organization={IEEE}
}

@inproceedings{Florence,
  title={Recognizing actions from depth cameras as weakly aligned multi-part bag-of-poses},
  author={Seidenari, Lorenzo and Varano, Vincenzo and Berretti, Stefano and Bimbo, Alberto and Pala, Pietro},
  booktitle={Proceedings of the IEEE conference on computer vision and pattern recognition workshops},
  pages={479--485},
  year={2013}
}

@inproceedings{SBU,
  title={Two-person interaction detection using body-pose features and multiple instance learning},
  author={Yun, Kiwon and Honorio, Jean and Chattopadhyay, Debaleena and Berg, Tamara L and Samaras, Dimitris},
  booktitle={2012 IEEE computer society conference on computer vision and pattern recognition workshops},
  pages={28--35},
  year={2012},
  organization={IEEE}
}

@inproceedings{IndRNN,
  title={Independently recurrent neural network (indrnn): Building a longer and deeper rnn},
  author={Li, Shuai and Li, Wanqing and Cook, Chris and Zhu, Ce and Gao, Yanbo},
  booktitle={Proceedings of the IEEE conference on computer vision and pattern recognition},
  pages={5457--5466},
  year={2018}
}

@article{BeyondJoint,
  title={Beyond joints: Learning representations from primitive geometries for skeleton-based action recognition and detection},
  author={Wang, Hongsong and Wang, Liang},
  journal={IEEE Transactions on Image Processing},
  volume={27},
  number={9},
  pages={4382--4394},
  year={2018},
  publisher={IEEE}
}

@inproceedings{SKCNN,
  title={Skeleton-based action recognition with convolutional neural networks},
  author={Li, Chao and Zhong, Qiaoyong and Xie, Di and Pu, Shiliang},
  booktitle={2017 IEEE international conference on multimedia \& expo workshops (ICMEW)},
  pages={597--600},
  year={2017},
  organization={IEEE}
}

@inproceedings{STGCN,
  title={Spatial temporal graph convolutional networks for skeleton-based action recognition},
  author={Yan, Sijie and Xiong, Yuanjun and Lin, Dahua},
  booktitle={Proceedings of the AAAI conference on artificial intelligence},
  volume={32},
  year={2018}
}

@article{EnsemNN,
  title={Ensemble one-dimensional convolution neural networks for skeleton-based action recognition},
  author={Xu, Yangyang and Cheng, Jun and Wang, Lei and Xia, Haiying and Liu, Feng and Tao, Dapeng},
  journal={IEEE Signal Processing Letters},
  volume={25},
  number={7},
  pages={1044--1048},
  year={2018},
  publisher={IEEE}
}

@article{MANs,
  title={Memory attention networks for skeleton-based action recognition},
  author={Li, Ce and Xie, Chunyu and Zhang, Baochang and Han, Jungong and Zhen, Xiantong and Chen, Jie},
  journal={IEEE Transactions on Neural Networks and Learning Systems},
  volume={33},
  number={9},
  pages={4800--4814},
  year={2021},
  publisher={IEEE}
}

@article{HCN,
  title={Co-occurrence feature learning from skeleton data for action recognition and detection with hierarchical aggregation},
  author={Li, Chao and Zhong, Qiaoyong and Xie, Di and Pu, Shiliang},
  journal={arXiv preprint arXiv:1804.06055},
  year={2018}
}

@article{T5,
  author  = {Colin Raffel and Noam Shazeer and Adam Roberts and Katherine Lee and Sharan Narang and Michael Matena and Yanqi Zhou and Wei Li and Peter J. Liu},
  title   = {Exploring the Limits of Transfer Learning with a Unified Text-to-Text Transformer},
  journal = {Journal of Machine Learning Research},
  year    = {2020},
  volume  = {21},
  number  = {140},
  pages   = {1-67},
  url     = {http://jmlr.org/papers/v21/20-074.html}
}

@article{luvizon2017learning,
  title={Learning features combination for human action recognition from skeleton sequences},
  author={Luvizon, Diogo Carbonera and Tabia, Hedi and Picard, David},
  journal={Pattern Recognition Letters},
  volume={99},
  pages={13--20},
  year={2017},
  publisher={Elsevier}
}

@inproceedings{vemulapalli2014human,
  title={Human action recognition by representing 3d skeletons as points in a lie group},
  author={Vemulapalli, Raviteja and Arrate, Felipe and Chellappa, Rama},
  booktitle={Proceedings of the IEEE conference on computer vision and pattern recognition},
  pages={588--595},
  year={2014}
}

@article{devanne20143,
  title={3-d human action recognition by shape analysis of motion trajectories on riemannian manifold},
  author={Devanne, Maxime and Wannous, Hazem and Berretti, Stefano and Pala, Pietro and Daoudi, Mohamed and Del Bimbo, Alberto},
  journal={IEEE transactions on cybernetics},
  volume={45},
  number={7},
  pages={1340--1352},
  year={2014},
  publisher={IEEE}
}

@inproceedings{seidenari2013recognizing,
  title={Recognizing actions from depth cameras as weakly aligned multi-part bag-of-poses},
  author={Seidenari, Lorenzo and Varano, Vincenzo and Berretti, Stefano and Bimbo, Alberto and Pala, Pietro},
  booktitle={Proceedings of the IEEE conference on computer vision and pattern recognition workshops},
  pages={479--485},
  year={2013}
}

@inproceedings{HON4D,
  title={Hon4d: Histogram of oriented 4d normals for activity recognition from depth sequences},
  author={Oreifej, Omar and Liu, Zicheng},
  booktitle={Proceedings of the IEEE conference on computer vision and pattern recognition},
  pages={716--723},
  year={2013}
}

@inproceedings{Rahmani,
  title={Real time action recognition using histograms of depth gradients and random decision forests},
  author={Rahmani, Hossein and Mahmood, Arif and Huynh, Du Q and Mian, Ajmal},
  booktitle={IEEE winter conference on applications of computer vision},
  pages={626--633},
  year={2014},
  organization={IEEE}
}

@inproceedings{Tran,
  title={Sparse spatio-temporal representation of joint shape-motion cues for human action recognition in depth sequences},
  author={Tran, Quang D and Ly, Ngoc Q},
  booktitle={The 2013 RIVF International Conference on Computing \& Communication Technologies-Research, Innovation, and Vision for Future (RIVF)},
  pages={253--258},
  year={2013},
  organization={IEEE}
}

@inproceedings{Xia,
  title={View invariant human action recognition using histograms of 3d joints},
  author={Xia, Lu and Chen, Chia-Chih and Aggarwal, Jake K},
  booktitle={2012 IEEE computer society conference on computer vision and pattern recognition workshops},
  pages={20--27},
  year={2012},
  organization={IEEE}
}

@inproceedings{Wang,
  title={Recognizing actions in 3d using action-snippets and activated simplices},
  author={Wang, Chunyu and Flynn, John and Wang, Yizhou and Yuille, Alan},
  booktitle={Proceedings of the AAAI Conference on Artificial Intelligence},
  volume={30},
  year={2016}
}
